\documentclass[twoside,11pt]{article}

\usepackage{jmlr2e}
\usepackage{amsmath}
\usepackage[ruled,vlined]{algorithm2e}
\usepackage[colorlinks,citecolor=blue,urlcolor=blue]{hyperref}
\usepackage{subfigure}
\usepackage{dsfont}
 \newcommand{\ind}{\raisebox{0.05em}{\rotatebox[origin=c]{90}{$\models$}}}
\usepackage{mathtools}

\usepackage{tikz}
\usepackage{float}
\usetikzlibrary{fit}
\tikzset{%
  highlight/.style={rectangle,rounded corners,fill=red!15,draw,
    fill opacity=0.5,thick,inner sep=0pt}
}
\newcommand{\tikzmark}[2]{\tikz[overlay,remember picture,
  baseline=(#1.base)] \node (#1) {#2};}
\newcommand{\Highlight}[1][submatrix]{%
    \tikz[overlay,remember picture]{
    \node[highlight,fit=(left.north west) (right.south east)] (#1) {};}
}

\SetKwInOut{Input}{$\triangleright$\ Input}
\SetKwInOut{Parameters}{$\triangleright$\ Parameters}
\SetKwInOut{Output}{$\triangleright$\ Output}
\SetKwComment{Comment}{$\triangleright$\ }{}

\newcommand{\mathbbm}[1]{\text{\usefont{U}{bbm}{m}{n}#1}} 
\SetCommentSty{comsty}

\def\Exp{{\ensuremath{\mathbb E}}}

\usepackage{nicematrix}
\usepackage{tabulary,booktabs}

\newcommand{\argmax}{\operatornamewithlimits{argmax}}

\newcommand{\spike}[2]
{\bgroup
  \sbox0{#2}%
  \rlap{\usebox0}%
  \hspace{0.5\wd0}%
  \makebox[0pt][c]{\rule[\dimexpr \ht0+1pt]{0.5pt}{#1}}
  \makebox[0pt][c]{\rule[\dimexpr -\dp0-#1-1pt]{0.5pt}{#1}}
  \hspace{0.5\wd0}%
\egroup}

\ShortHeadings{Multiclass ROC}{Liang Wang and Luis Carvalho}
\firstpageno{1}

\begin{document}

\title{Multiclass ROC}

\author{\name Liang Wang \email leonwang@bu.edu \\
       \addr Department of Mathematics and Statistics\\
       Boston University\\
       Boston, MA, 02215 USA
       \AND
       \name Luis Carvalho \email lecarval@math.bu.edu \\
       \addr Department of Mathematics and Statistics\\
       Boston University\\
       Boston, MA, 02215 USA}
       
\editor{}

\maketitle
\begin{abstract}%
Model evaluation is of crucial importance in modern statistics application. The construction of ROC and calculation of AUC have been widely used for binary classification evaluation. Recent research generalizing the ROC/AUC analysis to multi-class classification has problems in at least one of the four areas: 1. failure to provide sensible plots 2. being sensitive to imbalanced data 3. unable to specify mis-classification cost  and 4. unable to provide evaluation uncertainty quantification. Borrowing from a binomial matrix factorization model, we provide an evaluation metric summarizing the pair-wise multi-class True Positive Rate (TPR) and False Positive Rate (FPR) with one-dimensional vector representation. Visualization on the representation vector measures the relative speed of increment between TPR and FPR across all the classes pairs, which in turns provides a ROC plot for the multi-class counterpart. An integration over those factorized vector provides a binary AUC-equivalent summary on the classifier performance. Mis-clasification weights specification and bootstrapped confidence interval are also enabled to accommodate a variety of of evaluation criteria. To support our findings, we conducted extensive simulation studies and compared our method to the pair-wise averaged AUC statistics on benchmark datasets.
\end{abstract}

\section{Introduction}
\label{sec:intro}
For machine learning classification problems, the evaluation of a model is of crucial importance since it provides guidance on the choice of an ultimate models. In many fields of the evolving research, multi-class classification is of the interests. 
According to the its inputs, the evaluation metric can be generally divided into two categories. 
\subsection{Hard Classification Metric}
The first sets of naive evaluation metrics is based upon "hard classification" metrics where the input is typically how many number of observations are classified correctly/incorrectly:
\begin{itemize}
    \item True Positive(TP): classification result is \textbf{positive} and the ground truth is \textbf{positive}.
    \item False Positive(FP): classification result is \textbf{positive} but the ground truth is \textbf{negative}.
    \item True Negative(TN): classification result is \textbf{negative} and the ground truth is \textbf{negative}.
    \item False Negative(FN): classification result is \textbf{negative} and the ground truth is \textbf{positive}.
\end{itemize}
From which, one can define the True Positive Rate (TPR), True Negative Rate (TNR), False Positive Rate (FPR) and False Negative Rate Rate(FNR):
$$\text{TPR} = \frac{\text{Total True Positive }}{\text{Total Positive}}, \text{FPR} = \frac{\text{Total False Positive }}{\text{Total Negative}}$$
$$\text{TNR} = \frac{\text{Total True Negative }}{\text{Total Negative}}, \text{FNR} = \frac{\text{Total False Negative }}{\text{Total Positive}}$$
For a balanced classification problem, practitioners can highlight the model's capability to correctly identified either positive labels or the negative labels. With this criteria, the TPR and TNR are popularly adopted. Focusing on the percentage of correctly identified positive labels, the TPR is also termed as recall and sensitivity. Focusing on the percentage of correctly identified negative labels, the TNR is also termed as specificity. However, when we are dealing with imbalanced problem, focusing solely on the positive samples or the negative samples will mislead the evaluation to favour the majority classifier.

Under the imbalanced classification criteria, each of the above number captures part of the classifier performance, the commonly used classifier evaluations thus  usually consider a combination of the four sub-evaluation metrics with various weighting scheme.
For example, the Cohen’s Kappa \citep{cohen1960coefficient} and Matthews Correlation Coefficient (\textbf{MCC}) \citep{matthews1975comparison} can be considered respectively as the geometric mean and harmonic mean of 
$$\frac{TP\cdot TN - FP \cdot FN}{ (TP+FP) \cdot(FP +TN) }\quad  \text{and}\quad  \frac{TP\cdot TN - FP \cdot FN}{ (TP+FN) \cdot(FN +TN) }$$
The $F_1$ score can be considered as the harmonic mean of the precision and recall, i.e:
$$\frac{TP}{TP + FP} \quad \text{and}\quad \frac{TP}{TP + FN}$$

Empirical comparisons of those commonly used evaluation metrics for binary classification have been demonstrated in \citep{chicco2020advantages}, and \citep{chicco2021matthews}. With numerical examples, it is also concluded in those literature that MCC is more informative than Cohen's Kappa and Brier score. 

\citep{powers2010evaluation} also shown that the MCC can also be understood as the geometric mean of two regression coefficient of Markedness and Informedness.  \citep{gorodkin2004comparing} exploits this correlation coefficient relationship implied by binary MCC regression coefficients to generalize the MCC to a general mutli-class classification evaluation metrics. 
Although MCC can be considered as an ultimate choice of the "hard classification" metrics, it has a few shortcomings. The first one is that since the MCC is based upon the count number instead of the percentage rate,  its performance deteriorates seriously when the classification labels are imbalanced\citep{zhu2020performance}. Secondly, despite the fact that practitioner not only wants to know the classification result but also hope to understand how confident is the evaluation conclusion, the MCC statistics ignores the uncertainty behind the classification evaluation by simply counting the number of the incorrectness outputted by the model. Lastly, summarizing the classifier's performance with a single number overlooks many other useful information on the classifier that are available with visualization plot. For example, one might also want to get a sense of TPR and FPR performance with different level of confidence(threshold) from the evaluation as it is emphasized in \citep{drummond2006cost}.

\subsection{Soft Classification Metric}
Taking probabilities as the input to the metrics, a better alternative "Receiver Operating Characteristic (ROC) \citep{metz1978basic}" conducts judgement based on the "soft" classification result. The method not only provides the access to a comprehensive visualization against the random classifier but also circumvents the class skewness problem by directly taking the percentage rate (TPR and FPR) into consideration. If one wants a comprehensive score to summarize the model performance, the Area Under the Curve (AUC) score can be computed using numerical integration. From a statistics perspective, the AUC score has also been demonstrated to be equivalent to the probability of the event that a positive labeled observation will receive a higher probability ranking as it is compared to the negative labeled observation \citep{provost2000well}. This probability interpretation has embarked on a series of follow-up research in Mann-Whitney U-Statistic. 

Though remains the benchmark metric for modern binary classification problems, one disadvantage of the AUC method is its plausible equal assumption on the misclassification costs. In practice, it is of a rarity and thus often problematic to assume that nothing about the relative severity of misclassification is known. An alternative AUC taking advantage of the relative severity of the binary classification costs has been proposed in \citep{adams1999comparing}. Other methods such as \citep{provost1998robust} overcome this issue by constructing the ROC convex hull after the ROC analysis, through which the slope of isocost analysis can be applied to conduct the classifier evaluation. However, both methods dealing with the misclassification cost are within the binary classification setup.

Generalizing the AUC/ROC construction to a multi-class classification is challenging. To summarize some recent attempts, \citep{mossman1999three} and \citep{ferri2003volume} generalize the multi-class AUC through Volume Under the Surface (VUS) but the methods suffer from expensive computational budget and none of those methods can provide sensible visualization. Despite those drawbacks, there are generalizations of AUC statistics that received great popularity.
For example, \citep{provost2000well} approaches the generalization problem by taking a one-versus-all-wise average of the binary AUC. \citep{hand2001simple} adopt a similar average idea on the pair-wise binary AUC and demonstrated that the pair-wise AUC average preserves the class-skewness invariance property of the binary AUC. Employing the Mann-Whitney U-Statistic setup, \citep{ross2019aucmu} generalizes the AUC statistics through a partition definition of the order statistics. Although both \citep{ross2019aucmu}'s and \citep{provost2000well}'s methods are shown to be invariant to class-skewness, their methods discarded the ROC visualization and thus can only summarize the classifier performance with a scalar value for absolute comparison. Additionally, both of the methods lack of the capability of specifying the relative severity of misclassification costs, which is absolutely needed in practice to determine optimal classifier\citep{adams1999comparing}. Lastly, neither of those methods nor the binary AUC provides an uncertainty quantification on the evaluation conclusion. All those shortcomings substantiate a conclusion that better multi-class AUC method that is as well-accepted as the binary AUC method is yet to be discovered.
\subsection{Overall Contribution and Plans of the Paper}
Using a binomial DMF, we propose a new multi-class equivalent ROC and AUC for model evaluation. Our method borrows from the pair-wise AUC design \citep{provost2000well} by constructing a True Positive Rate Matrix $\mathbb{M}^{tp}$ and False Positive Rate Matrix $\mathbb{M}^{fp}$ among all pairs of the classes. Then we center the factorized components by taking the mean of the two factorized  matrices. Due to a flexible factorization weight, the factorization model also entitles the relative severity specification through factorization weights. Moreover, we demonstrate that the rank-one factorized components can be re-normalized to be of range $[0,1]$ and thus be interpreted as the shared threshold effects implied by the pair-wise averaged AUC.  The normalized vector provides not only a multi-class equivalent ROC plot but also an AUC score that summarizes the comprehensive performance of the classifiers. As a summary of the advantages of our factorization evaluation metrics over the existing multi-class evaluation metric:
\begin{itemize}
    \item The new metric provides access to a visualization plot that compares the classifier under different confidence(threshold) circumstances.
    \item The new metric is invariant to the label imbalance within the classification problem.
    \item The new metric enables the specification of relative severity for misclassification.
    \item The new metric quantifies the evaluation uncertainty with a resonable confidence interval
\end{itemize}

Our paper is organized as follow: In Section \ref{sec:background}, we provide some background information on the pair-wise multi-class AUC analysis.
In Section \ref{sec:model}, we formally introduce our evaluation metrics and provide some analyses on its desiring property. 
In Section \ref{sec:RocAlgo}, we summarize the implementation into an algorithm to facilitate applications. In Section {\ref{sec:RocResult}}, we provides both simulation and real-data examples to demonstrate the desired properties mentioned previously. In Section \ref{sec:end} we conclude the paper with a discussion and potential improvements.
\section{Backgrounds on Pair-wise AUC}
The pair-wise AUC design firstly connects the numerical integration to a ranking statistics and then generalizes the ranking idea toward a multi-class setup.
\label{sec:background}
\subsection{AUC and Mann Whitney U Statistics}
For a binary classification outcome with index $i$ and class label $Y_i$ taking binary value of $0$ or $1$, a classifier usually takes some covariates $X_i$ as input and outputs a simplex score vector $(S_i, 1-S_i)$. The score $S_i$ is usually taken to be a specific probability: $S_i \equiv \mathbb{P}_i \equiv \mathbb{P}(Y_i = 0)$, which represents the probability for the $i$-th observation to belong to class $0$. Given a specific probability threshold $t$, one can thus classify the $i$-th observation to class 0 if $\mathbf{S_i}$ is \textbf{greater than} the specified threshold $t$.  For an evaluation result with $n$ observations, the specifications of threshold $t$ will determine a "hard" classification assignment, which can then be used to compute evaluation metrics such as True Positive Rate (TPR denoted by $f(t)$) and a False Positive Rate (FPR denoted by $g(t)$). Both $f(t)$ and $g(t)$ are non-increasing functions with respect to threshold $t$, which implies $dg(t) \leq 0, df(t)\leq 0$ ($d$ here stands for differentiation operation). To summarize the performance of classifiers under different threshold circumstances, a ROC curve plots the TPR $f(t)$ against FPR $g(t)$ with a decreasing threshold $t$. 

 Based upon the constructed ROC curve, one can compute a more comprehensive score through numerical integration of $f(t)$ over $g(t)$ and obtain the Area Under the Curve (AUC). A perfect classifier will thus have a constant TPR $f(t) =1$ and increasing FPR $g(t)$ with respect to decreasing threshold $t$, giving an AUC value equal to 1. While a random classifier will have TPR $f(t)$ and FPR $g(t)$ increases at the same pace with respect to decreasing threshold $t$, giving an AUC value of 0.5. The relationship between $f(t)$ and $g(t)$ thus determine the quality of the classification model.

The AUC statistics has an interesting connection to Mann-Whitney U statistics. To observe the connection, denote $\{i\in I_k\}$ as the sets that the $i$-th observation belongs to class $k$, we could express the TPR and FPR with the following probability representation:
$$f(t) = \mathbb{P}(S_i> t | i\in I_0) =1- \mathbb{P}(S_i\leq t | i\in I_0)$$
$$g(t) =\mathbb{P}(S_i> t | i\in I_1) =1- \mathbb{P}(S_i\leq t | i\in I_1)$$
The AUC statistics as a numerical integration can then be written in the following form:
\begin{equation}
\label{eq:auc_stat}
\begin{aligned}
\text{A(0,1)}& = \int_{g(t) = 0}^{g(t)=1} f(t) dg(t)\\ 
&= \int_{g(t)=0}^{g(t)=1} \mathbb{P}(S_i\leq t | i\in I_1) d(1- \mathbb{P}(S_i\leq t | i\in I_0)) \\
&=  -\int_{t=1}^{t=0} \mathbb{P}(S_i\leq t | i\in I_1) \mathbb{P}(S_i= t | i\in I_0) dt \\
& = \mathbb{P}[ S_i 1_{ \{X_i \in I_1 \}} \leq S_i 1_{ \{X_i \in I_0\}}]
\end{aligned}
\end{equation}
Where the integration interval of the third equality is ranged from $[1,0]$ due to the fact $\frac{dg(t)}{dt}\leq 0$. The mathematical interpretation in Eq~\eqref{eq:auc_stat} provides us with two methods of computing the AUC statistics:
\begin{itemize}
    \item If we interpret the AUC statistics as the probability that the class 0 score of label 0 samples is higher than the class 0 score of label 1 samples, the AUC statistics can be estimated by using non-parametric Mann-Whitney U statistic. 
    \item If we instead model parametrically the relationship between $\mathbb{P}(S_i\leq t | i\in I_1)$ and $\mathbb{P}(S_i\leq t | i\in I_0)$, we could numerically integrate the AUC score with a fine numerical grid on threshold $t\in [0,1]$.
\end{itemize}
Specifically by following the first probability interpretation, 
we can calculate the AUC score using the Mann-Whitney U statistic. That is we can denote the probability of being class 0 of $n$ different samples using a vector $\vec{S} = (S_1, S_2, ....S_n)$. Then we can sort the score in an ascending order and obtain its order statistics $\vec{R} = (R_1, R_2, \cdots, R_n)$
with true class label $\vec{Y} = (Y_1, Y_2, \cdots, Y_n)$. For the $i$-th observation from class 0,  the number of points from class 1 with a score lower than the points from class $0$ is given by:
\begin{equation}
    U(0,1) = \sum_{i\in I_0} (R_i - i)
\end{equation}
Denote the number of points from class 0 $(|\{i\in I_0\}|)$ as $n_0$, the number of points from class 1 $(|\{i\in I_1\}|)$  as $n_1$. It is obvious to conclude that there are in total $n_0 \times n_1$ possible sample elements of statistics $U(0,1)$, which gives the following estimate on the probability of the ranking:
\begin{equation}
    \hat{A} \text{(0,1)} = \frac{U(0,1)}{n_0\times n_1}
    \label{eq:mann_auc}
\end{equation}
The AUC score $\text{A} (0,1)$ is indexed with (0,1) to emphasize that we are using the class 0 score (probability of the observation being class 0) to compare the observations from class 1 against the observations from class 0. For binary classification, the order of the index does not matter by having $A(0,1)$ = $A(1,0)$, but this equality  in general does not hold if we generalize to a multi-class classification problem. \citep{hand2001simple} adopted a pair-wise design to reduce the multi-class classification to multiple binary classifications.
\subsection{Generalizing AUC to Multiclass via Pair-wise Specification}
The design of the sub-binary problem is constructed by focusing only on two out of the multiple classification labels. To illustrate the pair-wise AUC concept for multi-class classification problem, we provide the following diagram:
\begin{figure}[htp]
\centering
\includegraphics[width=\textwidth]{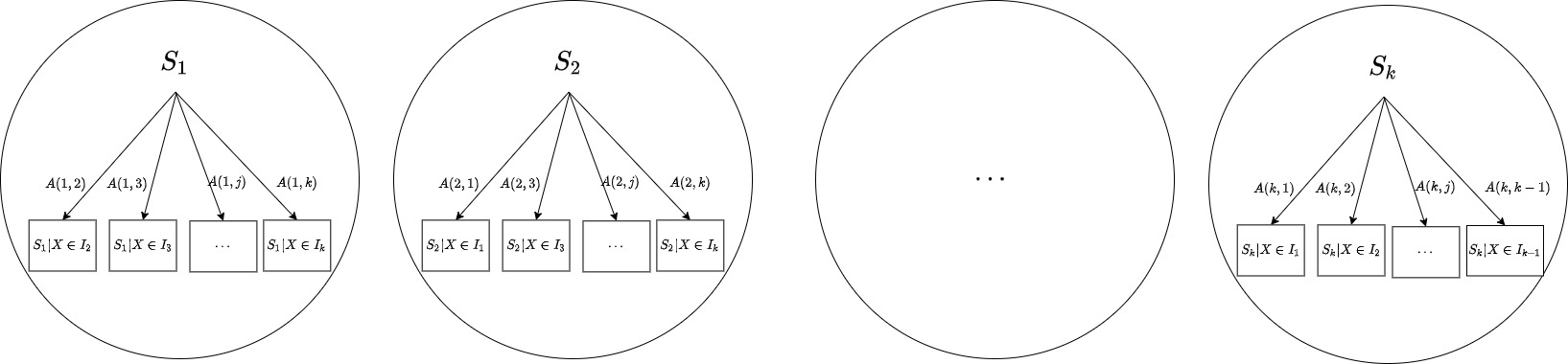}\hfill
\caption[Pair-wise AUC partition]{\textbf{Pair-wise AUC partition}}
\label{fig:partition}
\end{figure}
\\
As it is indicated in Figure \ref{fig:partition}, if $S_l$ is interpreted as classification probability for label $l$, we can potentially have $k$ different scores being defined with a simplex relationship $:\sum_{l=1}^k S_{l}= 1$. Then for each of the $k$ different scores, we can construct $k-1$ pairs of score comparison, concluding $K = k(k-1)$ pairs of the pair-wise AUC score for comparison.
Consequently, for each index $l$ with range $1,\ldots, k(k-1)$, it corresponds to a comparison index $(i,j), i\neq j, i,j\in [1,\ldots,k]$. Given a pair-wise index $l$ or equivalently the pair $(i,j)$, we can formally define the cardinality of class 0 and class 1 as:
\begin{equation}
        n_0(l) \equiv = n_0(i,j) = |\{ \text{obs} \in I_i\}| \quad n_1(l) \equiv n_1(i,j) = |\{ \text{obs} \in I_j\} | \\
    \label{eq:binarycard}
\end{equation}
where $\text{obs}$ stands for observations, $|\cdot|$ stands for the cardinality of a set, and $I_i$ stands for class $i$ assignment using the ground truth label. For each pair index $(i,j)$, its score can either be computed using numerical integration given in Eq~\eqref{eq:auc_stat} or using the Mann-Whitney U formula given in Eq~\eqref{eq:mann_auc}. \citep{hand2001simple} proposed the following score to measure the classifier performance:
\begin{equation}
\label{eq:provo2000}
    \begin{aligned}
    \hat{\mathcal{M}} = \frac{1}{k(k-1)} \sum_{i=1}^{k} \sum_{j\neq i, j =1}^k  \hat{A}(i,j)
    \end{aligned}
\end{equation}
Eq~\eqref{eq:provo2000} is invariant to class-skewness since it can be considered as a pair-wise average of all binary AUC statistics, which are themselves invariant to class skewness. Equivalently, the $\mathcal{M}$ statistics can be considered as an equal-weighted expectation of
$$\mathcal{M} = \Exp_{\{ (i,j) \in \mathcal{K}\}} [A(i,j)] = \Exp_{k} [ \int f_k(t) \Delta g_k(t) dt] = \Exp_k \Exp_t[f_k(t) \Delta g_k(t)]$$
The statistics implicitly assume that $f_k(t)$ and $g_k(t)$ are given deterministic numbers and thus provide a point-wise estimate. However, similar to the soft and hard classification from MCC to AUC or from k-means to the Gaussian mixture,  it is always of interest to investigate the behavior of the statistics if we generalize $f_k(t)$ and $g_k(t)$ to follow a random setup. This random generalization can additionally provide us with an uncertainty quantification through bootstrapped confidence interval.  For the binary evaluation, \citep{balaswamy2015confidence} imposed a Generalized Half Normal assumption on $f_k(t)$ and $g_k(t)$ and proposed a corresponding confidence interval to the AUC analysis. In the next section, we will explore the possibility of generalizing $f_k(t), g_k(t)$ to be random under binomial factorization setup.
\section{A Joint Binomial Factorization Model}
\label{sec:model}
In this section, we demonstrate that step-by-step how we construct our multi-class ROC curve and how to compute its corresponding AUC statistics.
\subsection{An Example of Binary Matrix Construction}
\label{subsec:BinaryExample}
Although we aim at a generalization toward multi-class ROC curve, it is intuitive for us to start with the description of the binary counterpart.
Given a decreasing threshold $\mathbb{T} = {\{t_1, t_2, \cdots t_T}\}$ and unitary space $\mathbb{Z} =[0,1]$, we start by computing the True Positive Rate(TPR) vector $\mathbb{M}^{tp}(0,1) \in \mathbb{Z}^{T\times 1}$ and False Positive Rate(FPR) vector $\mathbb{M}^{fp}(0,1) \in \mathbb{Z}^{T\times 1}$ by referencing the 1st class(class 0) as the positive class and the 2nd class(class 1) as the negative class:
\begin{equation}
  \mathbb{M}^{tp}(0,1) = \left(\begin{array}{*4{c}}
                        \tikzmark{left}{$f_{11}$} \\
                        f_{21} \\
                        \vdots\\
                        f_{T1} 
                        \end{array}\right)
\qquad
\mathbb{M}^{fp}(0,1) = \left(\begin{array}{*4{c}}
                        \tikzmark{left}{$g_{11}$} \\
                        g_{21}\\
                        \vdots \\
                        g_{T1}
                        \end{array}\right)
\end{equation}
In contract, if we reference the 2nd class(class 1) as the negative class and the 2nd class(class 1) as the positive class, we can compute another set of vectors $\mathbb{M}^{fp}(0,1) \in \mathbb{Z}^{T\times 1}$ and $\mathbb{M}^{tp}(0,1) \in \mathbb{Z}^{T\times 1}$:
\begin{equation}
  \mathbb{M}^{tp}(1,0) = \left(\begin{array}{*4{c}}
                        \tikzmark{left}{$f_{12}$} \\
                        f_{22} \\
                        \vdots\\
                        f_{T2} 
                        \end{array}\right)
\qquad
\mathbb{M}^{fp}(1,0) = \left(\begin{array}{*4{c}}
                        \tikzmark{left}{$g_{12}$} \\
                        g_{22}\\
                        \vdots \\
                        g_{T2}
                        \end{array}\right)
\end{equation}
Due to the label symmetry in binary classification problem, we in fact have 
\begin{equation}
    \begin{aligned}
        1  &= f_{t1} + g_{t2}, \forall t\in \mathbbm{T}\\
        1  &= g_{t1} + f_{t2}, \forall t\in \mathbbm{T}
    \end{aligned}
    \label{eq:binary_relationship}
\end{equation}
We can concatenate $\mathbb{M}^{tp}(0,1)$ and $\mathbb{M}^{tp}(1,1)$ together into matrices $\mathbb{M}^{tp} \in \mathbb{Z}^{T\times 2}$ and $\mathbb{M}^{fp} \in \mathbb{Z}^{T\times 2}$ and obtain:
\begin{equation}
\begin{aligned}
  \mathbb{M}^{tp} &=\bigg(
                       \mathbb{M}^{tp}(0,1), \mathbb{M}^{tp}(1,0)
                        \bigg)
                = \left(\begin{array}{*4{c}}
                        \tikzmark{left}{$f_{11}$} & f_{12} \\
                        f_{21} & f_{22} \\
                        \vdots & \vdots \\
                        f_{T1} & f_{T2}
                        \end{array}\right)
\\
\mathbb{M}^{fp} &=\bigg(
                       \mathbb{M}^{fp}(0,1), \mathbb{M}^{fp}(1,0)
                        \bigg)
                = \left(\begin{array}{*4{c}}
                        \tikzmark{left}{$g_{11}$} & g_{12} \\
                        g_{21} & g_{22} \\
                        \vdots & \vdots \\
                        \tikzmark{right}{$g_{T1}$} & g_{T2}
                        \end{array}\right)
\end{aligned}
\end{equation}
Due to this special relationship of binary classification problem in Eq~\eqref{eq:binary_relationship}, the ROC curve plots $f_{t1}$ against $g_{t1}$ to measure the model performance, which is equivalent to plotting $1-g_{t2}$ against $1-f_{t2}$. Hence, researchers have avoid the redundancy of computing two $FPR$ and $TPR$ by focusing on one choice of positive label assignment. This interesting property is however not preserved as we generalize toward the multi-class setup. Based upon $\mathbb{M}^{tp}$ and $\mathbb{M}^{fp}$, we can take either the first column vector or the second column vector of $\mathbb{M}^{tp}$ and $\mathbb{M}^{fp}$ and further concatenate them together row-wisely to obtain $\mathbb{M}$:
\[
\mathbb{M}_i \equiv 
\begin{pmatrix}
    \mathbb{M}_{*0}^{tp}     \\
    \mathbb{M}_{*0}^{fp}  
\end{pmatrix} 
\equiv
\bigg(
f_{11}, f_{21,}, \cdots, f_{T1}, g_{11}, g_{21}, \cdots, g_{T1}
\bigg)^\top \in \mathbb{Z}^{2T\times 1}
\]
and assume:
\[
\begin{pmatrix}
    \mathbb{M}_{i0}^{tp}     \\
    \mathbb{M}_{i1}^{fp}  
\end{pmatrix} \circ
\begin{pmatrix}
    n_1     \\
    n_0
\end{pmatrix}
\overset{ind}{\sim} \text{Bin} \bigg(\mu_{i} = 
\begin{pmatrix}
    m(\Lambda_{i0}^{tp})     \\
    m(\Lambda_{i0}^{fp})
\end{pmatrix}
,
\begin{pmatrix}
    n_1     \\
    n_0
\end{pmatrix}
\bigg)
\]
Such a model is saturated with $2T$ parameterization vector via $(\Lambda_{0}^{tp}, \Lambda_0^{fp})$, which provides the equivalence of $m(\Lambda_{i0}) = \mathbb{M}_{i0}$. The plotting of $m(\Lambda^{tp}_{i0})$ against $m(\Lambda_{i0}^{fp})$ is hence equivalent to the ROC plotting of $m(\Lambda_{i0}^{tp})$ against $m(\Lambda_{i0}^{fp})$.

\subsection{Construction of a Pair-wise Matrix}
\label{subsec:RocPairConstruct}
To generalize the binary case toward the multi-class set up, we operate on the same decreasing threshold $\mathbb{T} = {\{t_1, t_2, \cdots t_T}\}$ and binary space $\mathbb{Z} =[0,1]$. We start by building two matrices $\mathbb{M}^{tp} \in \mathbb{Z}^{T\times k(k-1)}$ and $\mathbb{M}^{fp} \in \mathbb{Z}^{T\times k(k-1)}$ where the two matrices store all pairs-wise (specified in Figure \ref{fig:partition}) TPR and FPR with respect to a decreasing threshold smoothed by the binary probability quantile $\mathbb{T}$. That is, for each pair-wise binary problem, we filter the rows the predicted probability matrix by focusing on observations whose ground truth label is either class $i$ or class $j$, based upon which we compute a smoothed threshold grids using the quantiles from the predicted probability matrix of those two classes:
\begin{equation}
  \mathbb{M}^{tp} = \left(\begin{array}{*4{c}}
                        \tikzmark{left}{$f_{11}$} & f_{12} & \cdots & f_{1K} \\
                        f_{21} & f_{22} & \cdots & f_{2K} \\
                        \vdots & \vdots & \ddots & \vdots \\
                        \tikzmark{right}{$f_{T1}$} & f_{T2} & \cdots & f_{TK}
                        \end{array}\right)
\Highlight[first]
\qquad
\mathbb{M}^{fp} = \left(\begin{array}{*4{c}}
                        \tikzmark{left}{$g_{11}$} & g_{12} & \cdots & g_{1K} \\
                        g_{21} & g_{22} & \cdots & g_{2K} \\
                        \vdots & \vdots & \ddots & \vdots \\
                        \tikzmark{right}{$g_{T1}$} & g_{T2} & \cdots & g_{TK}
                        \end{array}\right)
\Highlight[second]
\tikz[overlay,remember picture] {
  \draw[<->,thick,red,dashed] (first) -- (second) node [pos=0.66,above] {\scriptsize $\text{AUC}(0,1)$};
}
\label{eq:constructTPRFPR}
\end{equation}
Row-wisely, for a given $i = 1, 2, \cdots T$, $f_{i*}, g_{i*} \in [0,1]^{k(k-1)}$ correspond to the TPR and FPR given threshold $t_i$ across all the $k(k-1)$ pair-wise construction. Column-wisely, for a given $j = 1, 2, \cdots k(k-1) \equiv K$, $f_{*j}, g_{*j} \in [0,1]^{T}$  correspond to the $j$-th pair-wise TPR and FPR. $\text{AUC} (0,1)$ is thus a numerical integration of the first $\mathbb{M}^{tp}$ column with respect to the first $\mathbb{M}^{fp}$ column. Those two matrices contain valuable information for classification evaluation. To effectively summarize $\mathbb{M}^{tp}$ and $\mathbb{M}^{fp}$ using low dimensional vectors, we firstly emphasize the following observed properties of those two matrices:

\begin{itemize}
    \item Row-wisely, both $\mathbb{M}^{tp}$ and $\mathbb{M}^{fp}$ are monotonic in the sense that its value is increasing with respect to the decrease of the threshold $\mathbb{T}$.
    \item Row-wisely, $\mathbb{M}^{tp}$ and $\mathbb{M}^{fp}$ increases at different speeds. How fast does $\mathbb{M}^{tp}$ increases according to the increases $\mathbb{M}^{fp}$ determines the quality of the classifier. 
    \item Column-wisely, both $\mathbb{M}^{tp}$ and $\mathbb{M}^{fp}$ share the same pair-wise specification and thus should have similar column effects.
\end{itemize}
Due to this simple monotonic property of the $\mathbb{M}^{tp}$ and $\mathbb{M}^{fp}$, it is naturally hopeful to summarize the "speed of increment" across the pair-wise columns through an appropriately designed matrix factorization.
\subsection{The Factorization Model}
\label{subsec:RocPairModelSetup}
The Deviance Matrix Factorization(DMF)\citep{2021dmf} supports a binomial parametric assumption on the matrix. A follow-up research \citep{wang2024computational} studied the bayesian version of such a model with computational improvement. 
With $Z^{tp}$ and $Z^{fp}$ representing the count of true positive and the count of false positive, we can exploit 
the following data generating assumption to parameterize the multi-class ROC curve:
\begin{equation}
\label{eq:UW_DMF}
    \begin{aligned}
    Z^{tp}_{ij} \equiv \mathbb{M}_{ij}^{tp}W_{j}^{pos} & \overset{ind}{\sim} \text{Bin}(\mu^{tp}_{ij} = m(\eta^{tp}_{ij})| \eta_{ij}^{tp}= \Lambda_{i0}^{tp} + (\Lambda^{tp}_i)^{\top} V_j, W_{j}^{pos})\\ 
    Z^{fp}_{ij} \equiv \mathbb{M}_{ij}^{fp}W_j^{neg} &\overset{ind}{\sim} \text{Bin}(\mu^{fp}_{ij} = m(\eta^{fp}_{ij})| \eta_{ij}^{fp}= \Lambda_{i0}^{fp} + (\Lambda^{fp}_i)^{\top} V_j, W_{j}^{neg})
    \end{aligned}
\end{equation}
Concatenate those two matrices and abbreviate the notations with 
\[
Z \equiv \begin{pmatrix}
    \mathbb{Z}^{tp}     \\
    \mathbb{Z}^{fp}  
\end{pmatrix},
\mathbb{M} \equiv \begin{pmatrix}
    \mathbb{M}^{tp}     \\
    \mathbb{M}^{fp}  
\end{pmatrix}, \Lambda_0 \equiv \begin{pmatrix}
    \Lambda_0^{tp}     \\
    \Lambda_0^{fp}  
\end{pmatrix}, \Lambda \equiv \begin{pmatrix}
    \Lambda^{tp}     \\
    \Lambda^{fp}  
\end{pmatrix}, \eta \equiv \begin{pmatrix}
    \eta^{tp}     \\
    \eta^{fp}  
\end{pmatrix}
, W \equiv
 \begin{pmatrix}
    W^{pos}      \\
    W^{neg}
\end{pmatrix}
\]
The data generating process is equivalently:
$$Z_{ij} \equiv \mathbb{M}_{ij}W_{j} \overset{ind}{\sim} \text{Bin} (\mu_{ij} = m(\eta_{ij})| \eta_{ij} = \Lambda_{i0} + \Lambda_i^{\top} V_j, W_{j})$$
To appropriately account for the observations we made in Section~\ref{subsec:RocPairConstruct}, we impose the following constraint on the factorized components $\Lambda$ and $V$.
\begin{itemize}
    \item Row-wisely, The increment $\mathbb{M}^{tp}$ and $\mathbb{M}^{fp}$ can be decomposed into shared threshold effect (modeled by $\Lambda_{i0}$) and columns specific effect $\Lambda_i^\top V_j$. $\Lambda_{i0}$s are monotonic with respect to the decrease of the threshold and are independent(constructed to be orthogonal) of the columns-specific effect $\Lambda_i^\top V_j$.
    \item With factorization rank $q=1$, $\mathbb{M}^{tp}$ and $\mathbb{M}^{fp}$ increases at different speed with the implicit relationship $\eta_{i}^{tp} = \alpha_i + \beta_i \eta_{i}^{fp}$ where $\alpha_i =\Lambda_{i0}^{tp} -  \frac{\Lambda_i^{tp}}{\Lambda_i^{fp}} \Lambda_{i0}^{fp} $ and $\beta_i =\frac{\Lambda_i^{tp}}{\Lambda_i^{fp}} $
    \item Column-wisely, $\mathbb{M}^{tp}$ and $\mathbb{M}^{fp}$ share the same column effect by having the same column matrix $V$. But within the columns of $\mathbb{M}^{tp}$ and $\mathbb{M}^{fp}$, the increment should be independent.
\end{itemize}
Mathematically, this is equivalent to solving the low-rank representation $(\hat{\Lambda}, \hat{V})$ by maximizing the binomial likelihood $\mathcal{L}$ with link $g(\cdot)$ assumption:
\begin{equation}
\label{eq:dmfcentered2}
\argmax_{\substack{\Lambda_0 \in \mathbb{R}^{T},V^\top \mathbf{1}_K = 0\\
(\Lambda, V) \in \{\mathcal{V}_{T,K}(q)\}}}
W \circ \mathcal{L}(\mathbb{M}, \Lambda_0 \mathbf{1}_K^\top + \Lambda V^\top)= 
\argmax_{\substack{ \Lambda_0 \in \mathbb{R}^{T},  V^\top \mathbf{1}_K =0 \\
\Lambda \in \tilde{\mathcal{S}}_{T, q},
V \in \mathcal{S}_{K, q}}}
W \circ \mathcal{L}(\mathbb{M}, \Lambda_0 \mathbf{1}_K^\top + \Lambda V^\top),
\end{equation}
where  
\begin{enumerate}
\item[i)] $q$ is the reduced dimension and is chosen to be one to maintain $\eta_i^{tp} = \alpha_i + \beta_i \eta_i^{fp}$:
\item[ii)] $W$ is the binomial weight for the observation of true positive count $Z^{tp}$ and false positive count $Z^{fp}$.
\item[iii)] $V$ has orthogonal columns and are orthogonal to $1_K$:
$V^\top V = I_q, V^{\top} \mathbf{1}_K = 0$, i.e $V$ belongs to the centered Stiefel manifold
$\mathcal{S}_{T, q}$;
\item[iv)] $\Lambda$ has pairwise orthogonal columns, that is, $\Lambda = U D$
with $U \in \mathcal{S}_{T, q}$ and $D = \text{Diag}\{d_1, \ldots, d_q\}$ with
$d_1 \geq \cdots \geq d_q$ so that $\Lambda^\top \Lambda = D^2$. This
space for $\Lambda$ is denoted as $\tilde{\mathcal{S}}_{T, q}$.
\end{enumerate}
This model can thus be solved through a centering trick after obtaining a full rank factorization of $\Lambda \in \tilde{\mathcal{S}}_{T, K},V \in \mathcal{S}_{K, K}$ with weight $W_{ij}$. That is we center the factorized components through a projector matrix $H_0$ defined according to $V_0$ with $V_0$ being the vector of one.
\begin{equation}
    \begin{aligned}
    \Lambda V^\top & = \Lambda V^\top H_0 + \Lambda V^\top (I_K - H_0 )\\
    & =\underbrace{ \Lambda V^\top V_0 (V_0^\top V_0)^{-1}}_{\Lambda_0} V_0^\top + \underbrace{\Lambda V^\top (I_K - H_0 )}_{\tilde{\Lambda} \tilde{V}^{\top}}
    \end{aligned}
    \label{eq:dmfROCcenter}
\end{equation}
The reduced component $\Lambda_0, V_0$ are of rank one and are orthogonal to the newly constructed $\tilde{\Lambda} \tilde{V}^\top$ through the construction of projector $H_0$. In the case of $q=1$, we have $\Lambda_0 = \sum_{j=1}^K \frac{1}{K} \Lambda$, which aims at averaging the pair-wise TPR and FPR across $K$ different classes on the centered Stiefel manifold given by $(\mathbf{1}_K, V^{\top})$. $\Lambda_0$ thus has the interpretation as the averaged TPR/FPR across the $k(k-1)$ pair-wise groups. In the special case of $K =1$, we have precisely the binary ROC equivalence by having $V = \mathbb{1}$ as shown in Section~\ref{subsec:BinaryExample}. The model setup in fact is closely related to summary ROC analysis in the literature with details shown in Section~\ref{subsec:RocSROCConn}.
\subsection{Specifying Pair-wise Misclassification Cost}
\label{subsec:RocPairClassCost}
As demonstrated in  \citep{metz1978basic}, \citep{provost1998robust}, and \citep{adams1999comparing}, ignoring the misclassification cost can sometimes provide misleading results for model selection. 
In fact, it has been shown in the same references that an optimal classification threshold can be obtained only if the misclassification cost is specified. For the binary classification evaluation, the choice of optimal threshold on the ROC curve is formulated as an optimization problem given both the classification cost and the prior probability of class occurrence. Specifically, given the misclassification costs $C(0), C(1)$, the prior probability of class occurrence $\pi(0), \pi(1)$ and a threshold $t$, one can compute the probability of misclassifying class $k$ as $f_t(k)$ and define a threshold $t$ dependent evaluation loss function:
$$L_t = C(0)f_t(0) \pi(0) + C(1)f_t(1)\pi(1)$$
To minimize the loss with respect to threshold $t$, the slope defined by the iso-cost line $\frac{\pi(0) C(0)}{\pi(1)C(1)}$ is shown to be tangent to the optimal threshold $t$ on the ROC curve. Since only the tangent points are the possible optimal, the analysis is also termed ROC Convex Hull (ROCCH). However, for a pair-wise multi-class AUC setup, the convex hull analysis is hardly applicable due to the absence of an equivalent ROC curve.

Notice that by default, this optimization setup in Eq~\eqref{eq:dmfcentered2} weight the observation of $\mathbb{M}_{ij}^{pos}$ and $\mathbb{M}_{ij}^{neg}$ according to the cardinality of the positive and negative samples defined by the binary pair $W_{j}= n_0(j) \times n_1(j)$ where index $j$ is defined as the $j$-th pair-wise index as illustrated in Figure \ref{fig:partition}. In some of the existing literature, the cardinality of positive samples and negative samples are also referenced as the prior (occurrence) probability of the pair $(i,j)$. 

Although the default cardinality weighting is consistent with the intuition that $\mathbb{M}_{ij}^{pos}$ and $\mathbb{M}_{ij}^{neg}$ are observed with higher confidence due to large bernoulli sample size, this weighting scheme favours the majority classifier by putting more factorization weights on the majority binary pairs. To flexibly allow more intuitive misclassification cost, we introduce additional weight $Q_{ij}$ specification to the factorization, which changes the optimization setup from Eq~\eqref{eq:dmfcentered2} to Eq~\eqref{eq:dmfcentered3}:
\begin{equation}
\label{eq:dmfcentered3}
\hat{\Lambda} \hat{V}^\top (Q)= 
\argmax_{\substack{ \Lambda_0 \in \mathbb{R}^{T},  V^\top \mathbf{1}_K =0 \\
\Lambda \in \tilde{\mathcal{S}}_{T, q},
V \in \mathcal{S}_{K, q}}}
\sum_{i=1}^{2T}\sum_{j=1}^K Q_{ij} W_{ij}\mathcal{L}_{ij}(M, \Lambda_0 \mathbf{1}_K^\top + \Lambda V^\top),
\end{equation}
With this additionally factorization weight $Q_{ij}$, the evaluation metric can be specified as proportional to the relative severity of the misclassification. For example, one can specify $Q_{j} = \frac{1}{W_{j}}$ if the practitioners want the evaluation metric to be absolutely independent from label imbalance. This special choice of $Q_j = \frac{1}{W_j}$ is referenced as unweighted DMF-ROC later since it weights each binary pair equally. As for the difference between $Q_{ij}$ and $W_{ij}$, we emphasize that $Q_{ij}$ is introduced only in the optimization while $W_{ij}$ is assumed to appear in the data generating process of TPR/FPR.

This specification enables the third advantage of our AUC metrics because it circumvents the construction of multiple ROC convex hulls by directly taking the weights specification into TPR-like and FPR-like factorization.
As we show in Section~\ref{subsec:ResultWeight}, this weight specification $Q_{ij}$ flexibly allows misclassification cost to effectively affect the model evaluation results as we expected. 

\subsection{Connection to sROC} 
\label{subsec:RocSROCConn}
In fact, if we adopt a logit link function and a factorization rank $q = 1$, our factorization model can be considered as a multi-class generalization of the binary \textit{summary ROC}(sROC) analysis  \citep{moses1993combining, arends2008bivariate}, which has been additionally shown to connect with ordinal regression in \citep{tosteson1988general}.

The sROC basically transforms the original TPR and FPR to a linear continuous space through $\text{logit}$ link function and analyzes the incremental speed of TPR relative to FPR through a regression coefficient. For given threshold $i$ of a binary classification problem $j=K=1$, the sROC estimated $a, \beta$ according to the following setup:
\begin{equation}
\begin{aligned}
D_{ij} &= \text{logit}(\mathbb{M}_{ij}^{tp}) - \text{logit}(\mathbb{M}_{ij}^{fp})\\
E_{ij} &= \text{logit}(\mathbb{M}_{ij}^{tp}) + \text{logit}(\mathbb{M}_{ij}^{fp})\\
D_{ij} &= \alpha_j + \beta E_{ij} + e_{ij}
\end{aligned}    
\label{eq:sROC}
\end{equation}
where $e_{ij}$ is a random error and is orthogonal to the linear predictor. After estimating the coefficient $b$ based upon discretely observed $\mathbb{M}_{ij}^{tp}$ and $\mathbb{M}_{ij}^{fp}$, we can reconstruct $\mathbb{M}_{ij}^{tp}$ and $\mathbb{M}_{ij}^{fp}$ according to a finer numerical grid of $E_{ij}$ and thus construct a smoothed ROC curve.  Moreover, once the distribution of $e_{ij}$ is known and its variance parameter is estimated, one can construct multiple ROC curves and thus its confidence interval through a naive bootstrapping on estimated parameters. 

In \citep{reitsma2005bivariate}, it is emphasized that $\mathbb{M}_{ij}^{tp}$ and $\mathbb{M}_{ij}^{fp}$ are negatively correlated across index threshold $i$ because those two metrics share the same threshold specification. Lowering the cutoff value will then lead to more observation with a positive result, thereby increasing the number of true positives but also the number of false positive results. To take this potential negative correlation into consideration, a bivariate model is proposed for binary classification problem $j=K=1$:
\begin{equation}
    \begin{aligned}
 \begin{pmatrix}
\mu_{ij}^{tp}\\
\mu_{ij}^{fp}
\end{pmatrix}
& \overset{i.i.d}{\sim}
N \big(
\begin{pmatrix}
\mu_{j}^{tp}\\
\mu_{j}^{fp}
\end{pmatrix}, \Sigma_1   \big) 
\\
    \begin{pmatrix}
    \text{logit}(\mathbb{M}_{ij}^{tp}\big| \mu_{ij}^{tp})\\
    \text{logit}(\mathbb{M}_{ij}^{fp}\big| \mu_{ij}^{tp})\\
    \end{pmatrix}
    & \overset{i.i.d}{\sim} N \big(
\begin{pmatrix}
\mu_{ij}^{tp}\\
\mu_{ij}^{fp}
\end{pmatrix}, \Sigma_2   \big) \\
    \end{aligned}
    \label{eq:bivariateROC}
\end{equation}
where the parameter $\Sigma_1, \Sigma_2$ are used to model the negative correlation. Alternatively \citep{rutter2001hierarchical} parametrized the ROC curve via binomial random effect model:
\begin{equation}
    \begin{aligned}
    \text{logit} \begin{pmatrix}
\mu_{ij}^{tp}\\
\mu_{ij}^{fp}
\end{pmatrix}
& \sim
N \big(
\begin{pmatrix}
\eta_{\textcolor{red}{j}}^{tp}\\
\eta_{\textcolor{red}{j}}^{fp}
\end{pmatrix}, \Sigma_1   \big) 
\\
    \begin{pmatrix}
    \mathbb{M}_{ij}^{tp}\\
    \mathbb{M}_{ij}^{fp}\\
    \end{pmatrix} \circ 
\begin{pmatrix}
    W_{ij}^{pos}\\
    W_{ij}^{neg}\\
    \end{pmatrix} 
    & \sim \text{Bin} \big(
\begin{pmatrix}
\mu_{ij}^{tp}\\
\mu_{ij}^{fp}
\end{pmatrix}, 
\begin{pmatrix}
    W_{ij}^{pos}\\
    W_{ij}^{neg}\\
    \end{pmatrix} 
\big) 
    \end{aligned}
    \label{eq:bivariateBinomialROC}
\end{equation}
where the correlation between $\mathbb{M}_{ij}^{tp}$ and $\mathbb{M}_{ij}^{fp}$ are modeled through $\Sigma_1$ but the computation is more expensive as it is compared to the Gaussian assumption in Eq~\eqref{eq:bivariateROC}. The setup is more close to our model in Eq~\eqref{eq:UW_DMF}.

Notice that all the sROC introduced concentrates on binary classification problems. Hence the sROC parameterization models the correlation of the threshold effect row-wisely since there is no pair effect. If we generalize the sROC curve parameterization to multi-class classification problem, the column-wise pair-wise correlation should be far more important to model than the row-wise threshold effect. The intuition is that we have $K = k(k-1)$ columns composed of $k$ classes, which makes the ability to distinguish different binary pairs at the same threshold to be highly correlated. This observation implies an index change of the binomial ROC index from Eq~\eqref{eq:bivariateBinomialROC} to: 
\begin{equation}
    \begin{aligned}
    \text{logit} \begin{pmatrix}
\mu_{ij}^{tp}\\
\mu_{ij}^{fp}
\end{pmatrix}
& \sim
N \big(
\begin{pmatrix}
\eta_{\textcolor{red}{i}}^{tp}\\
\eta_{\textcolor{red}{i}}^{fp}
\end{pmatrix}, \Sigma_1   \big) 
\\
    \begin{pmatrix}
    \mathbb{M}_{ij}^{tp}\\
    \mathbb{M}_{ij}^{fp}\\
    \end{pmatrix} \circ 
\begin{pmatrix}
    W_{ij}^{pos}\\
    W_{ij}^{neg}\\
    \end{pmatrix} 
    & \sim \text{Bin} \big(
\begin{pmatrix}
\mu_{ij}^{tp}\\
\mu_{ij}^{fp}
\end{pmatrix}, 
\begin{pmatrix}
    W_{ij}^{pos}\\
    W_{ij}^{neg}\\
    \end{pmatrix} 
\big) , \forall j \in [K]
    \end{aligned}
    \label{eq:bivariateBinomialDMFROC}
\end{equation}
The setup however still suffers computationally, our factorization model setup in Eq~\eqref{eq:UW_DMF} can then be considered as an approximation to Eq~\eqref{eq:bivariateBinomialDMFROC} because
conditional on threshold $i$, the pair-wise $(j\in[1, \ldots, K])$ correlation is modeled through covariance structure $\Sigma_1$.

More specifically, Eq~\eqref{eq:UW_DMF} assumes a common $V_j$ for both $\eta_{ij}^{tp}$ and $\eta_{ij}^{fp}$:
\begin{align}
\text{logit} (\mu_{ij}^{tp}) &= \Lambda_{i0}^{tp} + (\Lambda^{tp}_i)^{\top} V_j\\    
\text{logit} (\mu_{ij}^{fp}) &= \Lambda_{i0}^{fp} + (\Lambda^{fp}_i)^{\top} V_j
\end{align}
with factorization rank $q =1$, we have scalar $\Lambda_i, V_j \in \mathbb{R}$, which implies:
\begin{equation}
    \begin{aligned}
\text{logit} (\mu_{ij}^{tp}) &= \underbrace{\frac{\Lambda_i^{tp}}{\Lambda_i^{fp}}}_{\beta_i} \text{logit} (\mu_{ij}^{fp}) + \underbrace{\Lambda_{i0}^{tp} -  \frac{\Lambda_i^{tp}}{\Lambda_i^{fp}} \Lambda_{i0}^{fp}}_{\alpha_i}, \quad \forall j \in [K] \\
\text{logit} (\mu_{i}^{tp}) &= \alpha_i + \beta_i^\top \text{logit} (\mu_{i}^{fp})
    \end{aligned}
\end{equation}
That is our setup implies a linear relationship between $\text{logit} (\mu_{ij}^{tp}) $ and $\text{logit} (\mu_{ij}^{fp})$ whose mean are modeled via $(\eta_i^{tp},  \eta_i^{fp})$ and whose covariance $\Sigma_1$ is modeled through parameter $\Lambda_i, V^\top$.

\subsection{Visualization and Confidence Interval}
\label{subsec:RocConfid}
With the interpretation in Section~\ref{subsec:RocSROCConn}, we know that how fast does $\Lambda_0^{tp}$ increase with respect to $\Lambda_0^{fp}$
 represent the quality of the classifier. As a result, after summarizing $\mathbb{M}^{tp}$ and $\mathbb{M}^{fp}$ through a binomial rank one representation, we could simply plot the center of the factorized components, i. e, plot $\Lambda_0^{tp}$ against  $\Lambda_0^{fp}$ to visually inspect how fast on average does the row of $\mathbb{M}^{tp}$ increases with respect to the row of $\mathbb{M}^{fp}$.
To mimic the ROC plot in the binary classification problem, we further normalize the factorized components $\Lambda_0^{tp}$ and $\Lambda_0^{tp}$ through a monotonic link function $m(\cdot) = \text{logit}^{-1}(\cdot)$ to provide a multi-class equivalent ROC plot. 

The plot thus naturally determines the quality of the classifier with extensive empirical evidence given in Section \ref{sec:RocResult}. Like the construction of the AUC statistics, we could also summarize the classifier's performance with statistics $\mathcal{D}$ through integration on the centered components:
\begin{equation}
    \begin{aligned}
        \mathcal{D} \equiv \int \text{logit}^{-1} (\Lambda_0^{tp}) d \text{logit}^{-1}(\Lambda_0^{fp})    
    \end{aligned}
    \label{eq:Dstat}
\end{equation}
Additionally, since $\mathbb{M}_{ij} \overset{ind}{\sim} \text{Bin} (\mu_{ij} = m(\eta_{ij})| \eta_{ij} = \Lambda_{i0} + \Lambda_i V_j^{\top}, W_{ij})$ is implicitly assumed from the factorization,  we can simulate $\mathbb{M}^{tp}$ and $\mathbb{M}^{fp}$ after obtaining an estimation of $\eta^{tp}$ and $\eta^{fp}$ with weight $W_j$. The simulation naturally provides the access to a confidence interval around the ROC curve and thus the confidence interval of $\mathcal{D}$ statistics.

\section{Algorithm ROC-DMF}
\label{sec:RocAlgo}
Since the analysis in the previous section indicates a solid statistical connection to sROC using a factorization rank of 1 $(q=1)$, we provide efficient vectorization 
considering the normal equation reduces to a scalar update when $q= 1$. 

The factorization is conducted by concatenating the matrix
$\mathbb{M} = 
\begin{pmatrix}
    \mathbb{M}^{tp}\\
    \mathbb{M}^{fp}
\end{pmatrix}$
with $\mathbb{M} \in [0,1]^{2T \times k(k-1)}$. The factorized latent space is of the same dimension with $\eta \in \mathbb{R}^{2T \times k(k-1)}$. The factorized components $\Lambda$ is a vector of dimension $2T$: 
$\Lambda \in \tilde{\mathcal{S}}_{2T,1}$ and the basis matrix $V$ is also a vector of dimension $k(k-1)$: $V\in \mathcal{S}_{k(k-1),1}$. We denote the $t$-th scalar of $\Lambda$ as $\lambda_t$ and the $j$-th
scalar of $V$ as $v_j$;  $\forall i = 1,\cdots, 2T, j = 1, \cdots, k(k-1)$. With those notations, we can iterate the $\lambda$ and $v$ update by solving the following system of the equation:
\begin{equation}
    \begin{aligned}
        V^\top D_{i\cdot} V \lambda_i^{(t+1)} &= V^\top D_{i\cdot}
(V\lambda_i^{(t)} + D_{i\cdot}^{-1} G_{i\cdot}) \doteq
V^\top D_{i\cdot} Z_{i\cdot}^{(t)} \\
\Lambda^\top D_{\cdot j} \Lambda v_j^{(t+1)} &=
\Lambda^\top D_{\cdot j} (\Lambda v_j^{(t)} + D_{\cdot j}^{-1} G_{\cdot j})
\doteq \Lambda^\top D_{\cdot j} Z_{\cdot j}^{(t)},
    \end{aligned}
    \label{eq:DMFROCUpdate}
\end{equation}
where $S_{ij}, G_{ij}$ and $Z_{ij}$ are defined as:
\begin{itemize}
    \item  $W_{j} = n_0(j) \times n_1(j) \times Q_j$, $\eta_{ij} = \Lambda_i V_j$
    \item  $g(\cdot) = \text{logit}(\cdot)$, $\mathcal{V}(\mu_{ij}) = \mu_{ij} (1-\mu_{ij})$, $\mu_{ij} = g^{-1}(\eta_{ij})$
    \item $S_{ij} = w_{ij} \frac{{{g^{-1}}'(\eta_{ij})}^2}{\mathcal{V}(\mu_{ij})} $ and  $G_{ij} = \frac{{g^{-1}}'(\eta_{ij})}{\mathcal{V}(\mu_{ij})} w_{ij} (X_{ij} - \mu_{ij}) $
    \item $D_{i\cdot} \doteq \text{Diag}\{S_{i\cdot}^{(t)}\}$ with $S_{i\cdot}$
denotes the $i$-th row of $S$. Similarly, $\text{Diag}\{S_{\cdot j}^{(t)}\}$ with $S_{\cdot j}$
denotes the $j$-th column of $S$.
    \item $Z^{(t)}$ is the working response:
    \begin{equation}
Z_{ij}^{(t)} = \eta_{ij}^{(t)} + \frac{G_{ij}^{(t)}}{S_{ij}^{(t)}} =
\eta_{ij}^{(t)} + \frac{X_{ij} - \mu_{ij}^{(t)}}{{g^{-1}}'(\eta_{ij}^{(t)})}.
\end{equation}
\end{itemize}
Vectorizing Eq \eqref{eq:DMFROCUpdate} to obtain the following definition:
\begin{equation}
    \begin{aligned}
        \underline{V} &= \sqrt{S} \circ 
        (\begin{bmatrix}
            \underbrace{V, \cdots, V}_{2T\ \text{times}} 
        \end{bmatrix})^\top \\
        \underline{\Lambda} &= \sqrt{S} \circ 
        \begin{bmatrix}
            \underbrace{\Lambda ,
            \cdots, 
            \Lambda }_{k(k-1)\ \text{times}} 
        \end{bmatrix} \\
        \underline{Z} & = Z \circ S
    \end{aligned}
    \label{eq:DMFROC_prepare}
\end{equation}
$\underline{V}$, $\underline{\Lambda}$ and $\underline{Z}$ are matrix of dimension $2T \times k(k-1)$. With element-wise division $\oslash$ and square $(\cdot)^2$, the updates become:
\begin{equation}
    \begin{aligned}
        \Lambda^{(t+1)} &= V^\top \underline{Z}^\top \oslash \mathbf{1}_p^\top \underline{V}^2  \\
        V^{(t+1)} &= \Lambda^\top \underline{Z} \oslash \mathbf{1}_n^\top \underline{\Lambda}^2  
    \end{aligned}
    \label{eq:dmfROC_update}
\end{equation}
To facilitate the implementation, we summarize both the construction of the matrix $\mathbb{M}$ and the optimization steps in Algorithm \ref{Algo:Mauc}.
\\
\begin{algorithm}[H]
 \textbf{Definition:}
 Number of classes $k$, number of classification instance $n$.
 
 \textbf{Input:}
Classification probability $\mathbb{P} \in [0,1] ^{n\times k}$, factorization weight $Q_j = \mathbb{R}_+^{K}$, true class label $Y \in [1,\ldots, k]^{n}$, threshold quantile  $\mathbb{T} = \{\tau_1, \ldots, \tau_T\}$ with $0 \leq \tau_1 < \ldots \tau_T \leq 1$.

 \For(\hfill $\triangleright$ construct $\mathbb{M}^{tp}, \mathbb{M}^{fp}$ according to Eq~\eqref{eq:constructTPRFPR}){$i=1, 2, \ldots, k$ }{
    Extract probability vector from the $i$-th column of $\mathbb{P}$: $\mathbb{P}_i \equiv \mathbb{P}(Y = i) =  \mathbb{P}_{*i}$\\
    Compute the pair-wise threshold according to quantile $\mathbb{T}$ of $\mathbb{P}_i$.\\
    Define class $0$ samples $Y(0) = \{Y_l: Y_l \in I_i \}_{l=1}^n$\\
    \For{$j =1, 2, \ldots, k, j \neq i$ }{
        Define class $1$ samples $Y(1) = \{Y_l: Y_l\in I_j \}_{l=1}^n$\\
    \For{$t=1, 2, \ldots, T$ }{
    Compute the $\mathbb{M}^{tp}$ element according to $\tau_t$ quantile of $\mathbb{P}_i$\\
    Compute the $\mathbb{M}^{fp}$ element according to $\tau_t$ quantile of $\mathbb{P}_i$
    }
    Set $W_j = n_0(j) \times n_1(j)$ as in Eq~\eqref{eq:binarycard}.
  }
 }
\textbf{Factorization} on $ \mathbb{M} = [\mathbb{M}^{tp}, \mathbb{M}^{fp}]^\top$ wth weight $W_j = W_j\times Q_j$:
    
\For(\hfill $\triangleright$ initialization){$i = 1, \ldots, 2T$ and $j = 1, \cdots, k(k-1)$}{
  Set $\mu_{ij}^{(0)}$ as a perturbed version of $\mathbb{M}_{ij}$ and
  $\eta_{ij}^{(0)} = g(\mu_{ij}^{(0)})$\;
}
Set $L^{(0)}$ as the first $q$ columns of the $j$ index
$[\eta_{ij}^{(0)}]_{i=1,\ldots,2T; j=1,\ldots,k(k-1)}^\top$\;
\For(\hfill $\triangleright$ vectorized updates){$t = 0, 1, \ldots$ (until convergence) }{
    Compute $\underline{\Lambda}, \underline{V}, \underline{Z}$ as in~Eq~\eqref{eq:DMFROC_prepare}\;
    Update  $V$ and  $\Lambda$ according to Eq~\eqref{eq:DMFROCUpdate}\;
} 
Obtain $\Lambda_0^{tp}, \Lambda_0^{fp}$ through the projection on ($\hat{\Lambda}, \hat{V}^{\top}$) DNF result using Eq~\eqref{eq:dmfROCcenter}\\  
 \KwResult{Binomial factorized $\Lambda_0^{tp}$ and $\Lambda_0^{fp}$ and metric $\mathcal{D}$ according to Eq~\eqref{eq:Dstat}}
 \caption{\textbf{Muti-class AUC}}
 \label{Algo:Mauc}
\end{algorithm}

\section{Experiments}
\label{sec:RocResult}
To support our findings, we conducted simulation studies to demonstrate that 1. our method can successfully distinguish the classifiers of different performance quality, 2. our method is invariant to class skewness, 3. our ROC plot can reasonably take the weights specification into consideration, and 4. our ROC plot can provide reasonable confidence interval to quantify the evaluation uncertainty.
\subsection{Discriminative Experiment}
\label{subsec:RocExpDiscre}
To define a well-understood multi-class classification problem, we simulated multinomial regression data conditional on known covariates $X \in \mathbb{R}^{n\times p}$ and known coefficients $B \in \mathbb{R}^{p\times k}$. 
Specifically, the covariate $X$ and coefficients $B$ are indexed with sample size $n = 50,000$ and dimension $p = 10$. Their values are fixed after a row-wise/column-wise random sample from $X_i \sim N(0, I_p), B_{j} \sim N(1, I_p), \forall i = 1,\ldots n,  \forall j = 1, \ldots k-1.$  The number of class $k$ is chosen to be $5$ as a starting example. The classification assignment probability is obtained conditional on multinomial regression coefficient $B\in \mathbb{R}^{p\times (k-1)}$:
$$\vec{\mathbb{P}}_i = \big(\mathbb{P}(Y_i=1), \mathbb{P}(Y_i=2), \cdots, \mathbb{P}(Y_i=k) \big) = \text{Softmax}( (1, X_iB)) $$
The ground truth label $Y_i$ can then be obtained according to a multinomial random sample with probability $\vec{\mathbb{P}}_i$:
\begin{equation}
Y_i \sim  \text{Multinomial}( \vec{\mathbb{P}}_i)
\label{eq:multi_sample}
\end{equation}
Or to make the setup even simpler, we can also simply assign the label according to the highest element of simulated vector $\vec{\mathbb{P}}_i$:
\begin{equation}
Y_i = \argmax_{i = 1,\ldots, k} \vec{\mathbb{P}}_i
\label{eq:multi_deter_sample}
\end{equation}
Taking $\vec{\mathbb{P}}_i$ as the output of a classification model, the label setup in \eqref{eq:multi_deter_sample} thus corresponds to
 a perfect classifier since we are essentially fitting the label according to the probability assignment. The label setup in \eqref{eq:multi_sample} will correspond to a more realistic classifier due to the random multinomial assignment from $\vec{\mathbb{P}}_i$.
 
 With fixed random seed, the number of observations associated with each class from Eq~\eqref{eq:multi_sample} is correspondingly 9,534, 6,042, 11,586, 12,957, and 9,881. The number of observations of each class from Eq~\eqref{eq:multi_deter_sample} is correspondingly 9471, 5,111, 12,159, 13,543, 9,716. Below, we plotted the histogram of the generated class label:
\begin{figure}[H]
\centering
\subfigure[Random Label]{\includegraphics[width=0.4\textwidth]{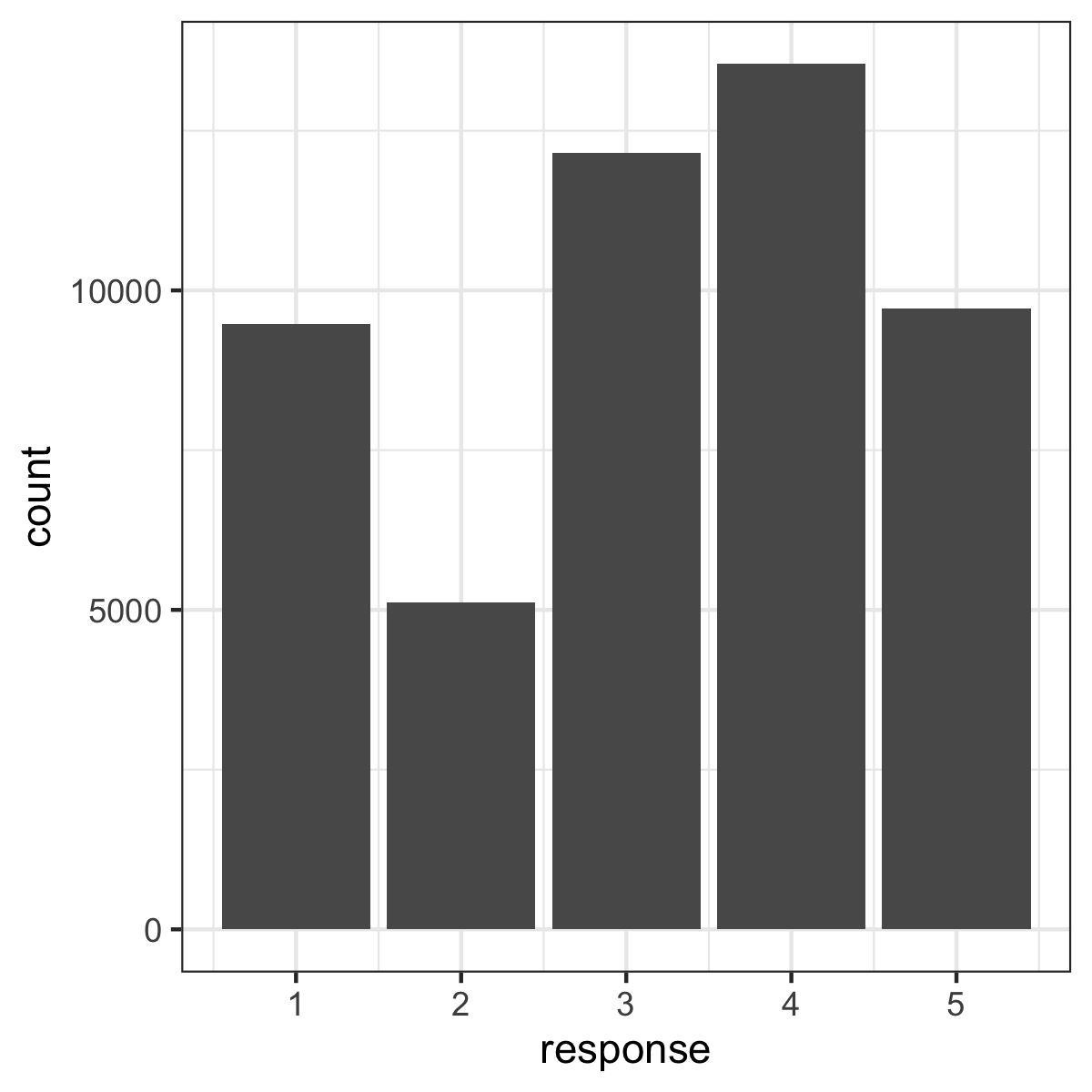}}
\subfigure[Deterministic Label]{\includegraphics[width=0.4\textwidth]{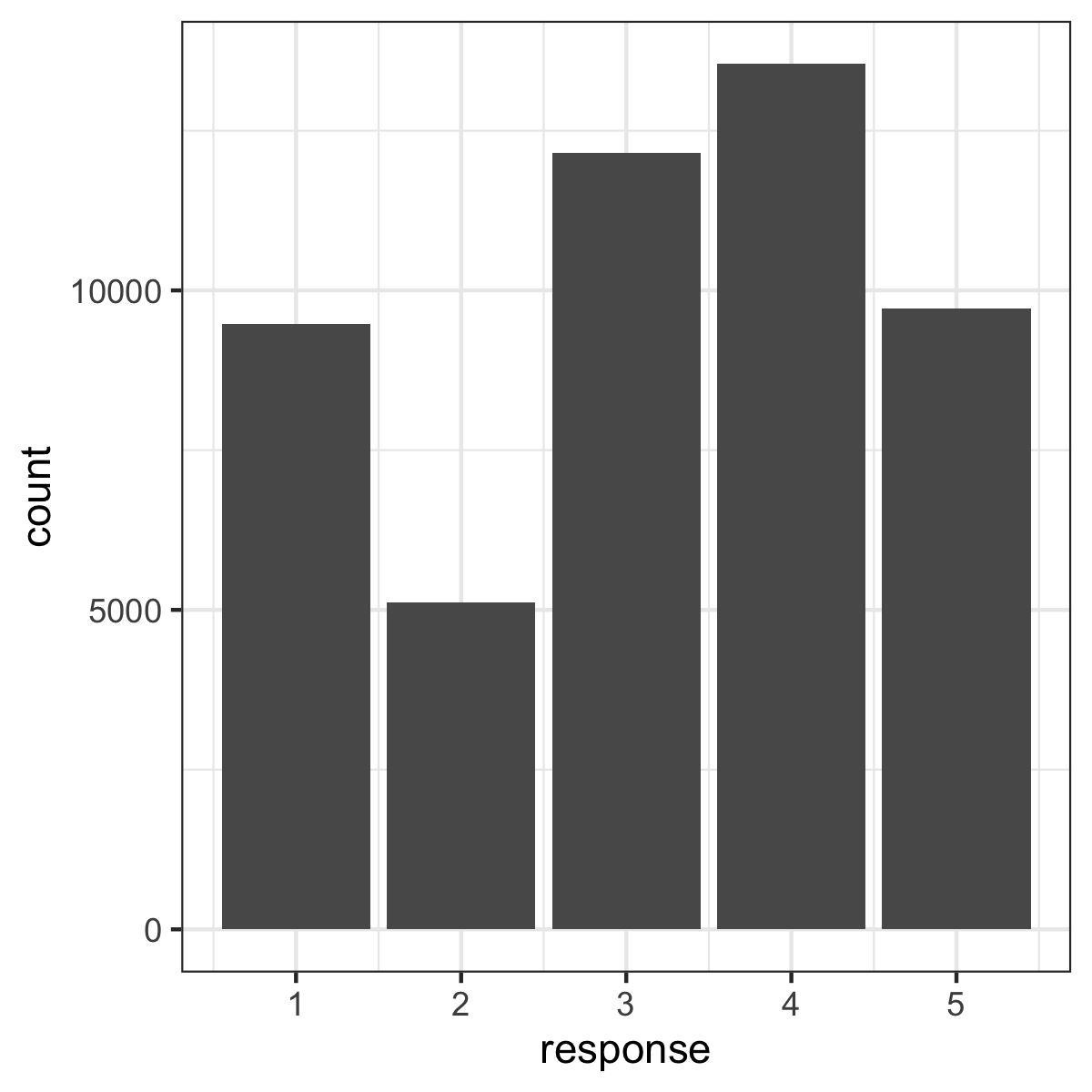}}
\caption[Hisogram of simulated class labels]{\textbf{{Hisogram of simulated class labels}}}
\label{fig:simu_hist}
\end{figure}
Then to create classifiers of different performance quality, we train classifies with partial availability of the generating covariance $X$.
That is, we gradually decrease the availability of the dimension of covariate $X$ to obtain classifiers with inferior prediction capability. Specifically, we fit a multinomial regression with only the first $d(d = 1, 2, \cdots p)$ columns of $X$, denoted as $X^{d}$. 
\begin{equation*}
    X  =  
\begin{bmatrix}
    \underbrace{\spike{30pt}{$X_{*1}$} ,\ldots, \spike{30pt}{$X_{*d}$}}_{X^d}, \ldots, \spike{30pt}{$X_{*p}$} 
\end{bmatrix}
\end{equation*}
To benchmark different levels' classifiers against the random classifier, we also provide one classifier with another independent simulation of $\tilde{X}_i \sim N(0, I_p), i = 1,2,\ldots, n $. Predicting the outcome conditional on $\tilde{X}$ is considered as a random classifier because we know $\tilde{X} \ind Y$ from the generating process.

Lastly, for each of the fitted multinomial regression conditional on $X^d$, we can obtain a fitted probability $\hat{\mathbb{P}}^d_i = \mathbb{P}(Y_i|X_i^d B^d) \in [0,1]^k$  with $B^d \in \mathbb{R}^{d\times k}$. We then input both the true class label $Y$ and the fitted probability $\hat{\mathbb{P}}^d$ into our evaluation Algorithm \ref{Algo:Mauc} with $Q_j = \frac{1}{n_0\times n_1}$ to weight each binary pair equally. From the output the algorithm, we can easily plot $\text{logit}^{-1}(\Lambda_0^{tp})$ against $\text{logit}^{-1}(\Lambda_0^{tp})$ to construct the ROC curves. The constructed curves are provided below:
\begin{figure}[H]
\centering
\subfigure[Random Label]{\includegraphics[width=0.49\textwidth]{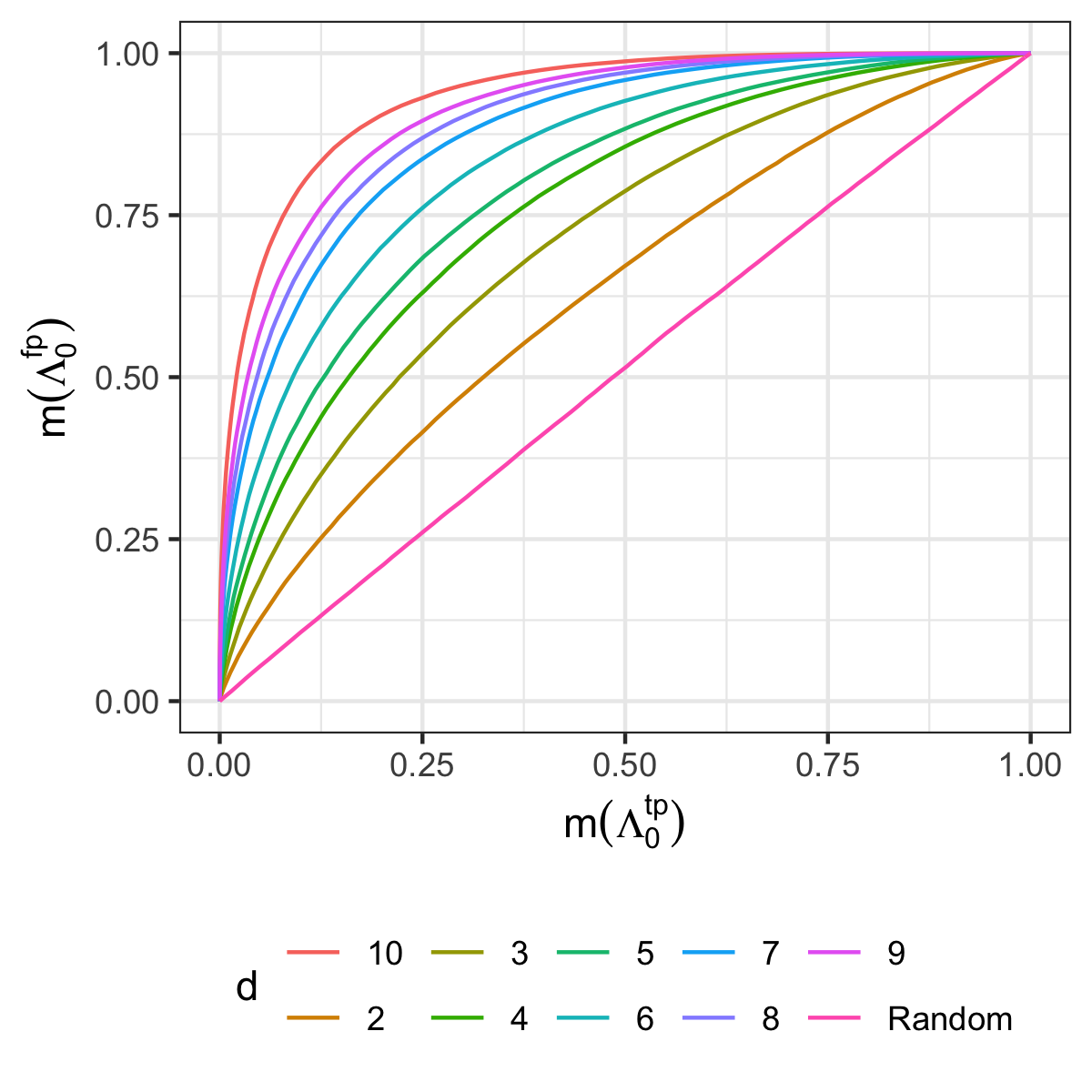}}
\subfigure[Deterministic Label]{\includegraphics[width=0.49\textwidth]{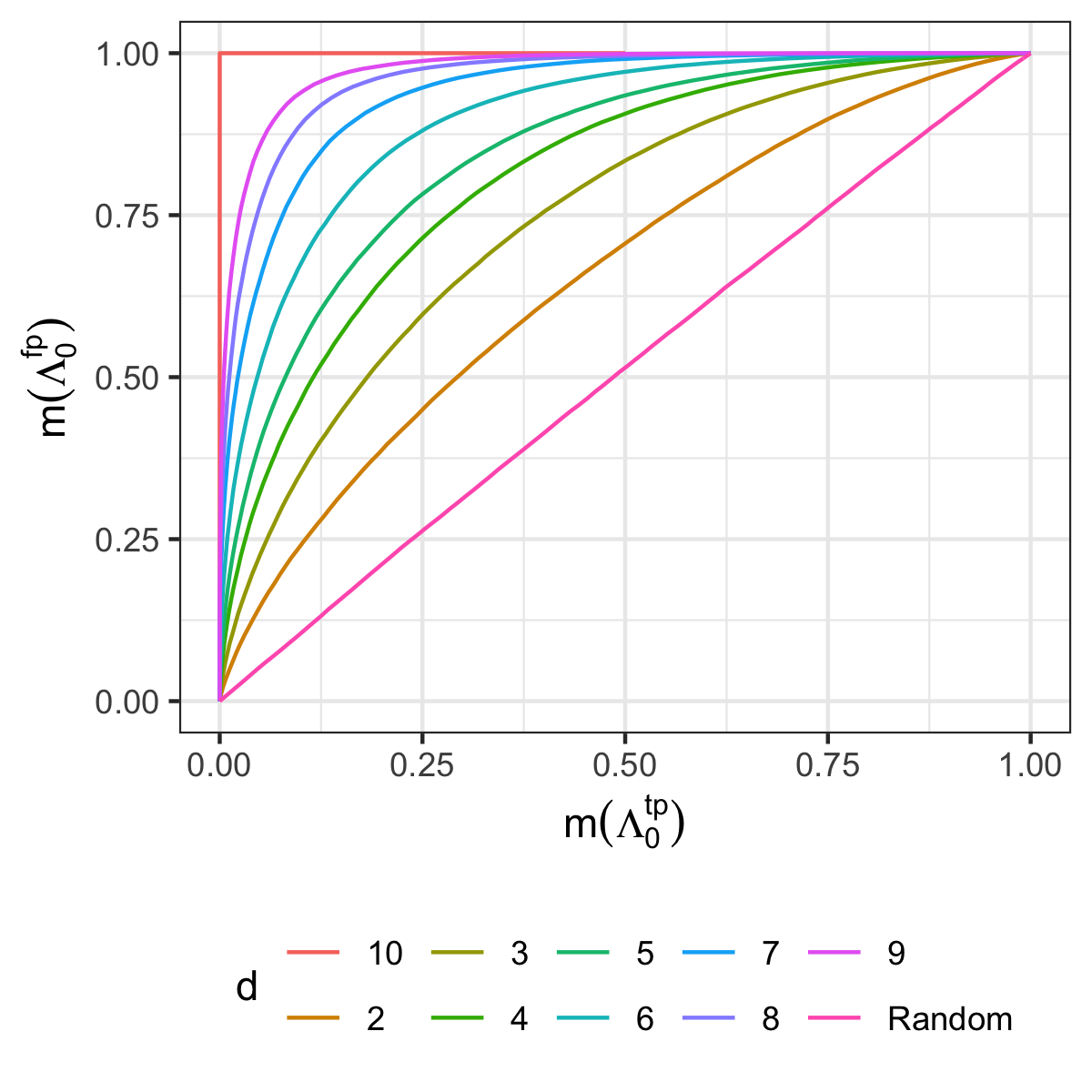}}
\caption[Discriminative experiment]{\textbf{Discriminative experiment}}
\label{fig:discre}
\end{figure}
As we can see from Figure \ref{fig:discre}, the optimal classifier with the full availability of $X^{p}$ is plotted as the top-left curve with the highest AUC value.  The classifier's performance deteriorates as we gradually decrease the availability of the covariate information defined by parameter $d$. 
The ROC curve of deterministic label (Eq~\eqref{eq:multi_deter_sample}) is higher than the one of sampling labels (Eq~\eqref{eq:multi_sample}) with the best performance ROC = 1 given the full availability of predictor.
The result provides the first evidence that our DMF factorization provides effective discriminative ROC curve for multi-class classification. To further investigate the invariance to class-skewness property, we provide more empirical examples in the next subsection.
\subsection{Class-skewness Experiment}
\label{subsec:ResultClassSkew}
From the simulation setup, we could obtain an imbalanced classification problem defined by $(Y_n, X_n)$. To make the classification problem more realistic, we take the random label assignment of Eq~\eqref{eq:multi_sample} as the groud truth label.
Then to create a simulation example with a balanced class assignment, we simply sample an equal number $n_{sub} = 6,042$ of observations within each class. Based upon this balanced simulation $Y_{n_{sub}}, X_{n_{sub}}$, we further create different level of skewness by sampling disproportionately the observations from $k$ classes according to a simplex vector $\vec{w}\equiv (w_1, w_2,\ldots, w_k)$. 

The vector $\vec{w}$ is repeatedly simulated from Dirichlet distribution $\vec{w} \sim \text{Dir}(\vec{\alpha})$ with parameter $\vec{\alpha} \equiv (\alpha_1, \alpha_2, \ldots, \alpha_k)$. For simplicity, we choose $\alpha_1 = \alpha_2 = \ldots = \alpha_k = \alpha$. To investigate the impact of parameter $\alpha$ on simulated skewness, we quantify the data skewness using the ratios among the Dirichlet-sampled weights $\frac{w_i}{w_j}, \forall i \neq j$. To summarize the skewness of the $k$ different classes, we denote random variable $Z = \max_{i\neq j}\frac{w_i}{w_j}$ as the surrogate of the overall class-skewness. The random variable $Z$ can be understood as the ratio between the cardinality of majority class and the cardinality of minority class. As we can conclude from the setup, the higher the $Z$ value is, the higher the class skewness is within the experiment dataset. We also empirically estimate and visualize (in Figure~\ref{fig:Dir_visu}) the tail probability $P(Z>c|\alpha)$ with 10,000 independent Dirichlet random sampling:
$$\hat{P} (Z>c|\alpha) = \frac{\sum_{i=1}^{10,000} 1_{\{Z_i>c\}}}{10,000}$$
To investigate the impact of parameter $\alpha$ on level of class skewness under different number of classes $(k)$, we also calculated the probability estimate with number of classes $k=10$:
\begin{figure}[h]
\centering
\subfigure[k=5]{\includegraphics[width=0.49\textwidth]{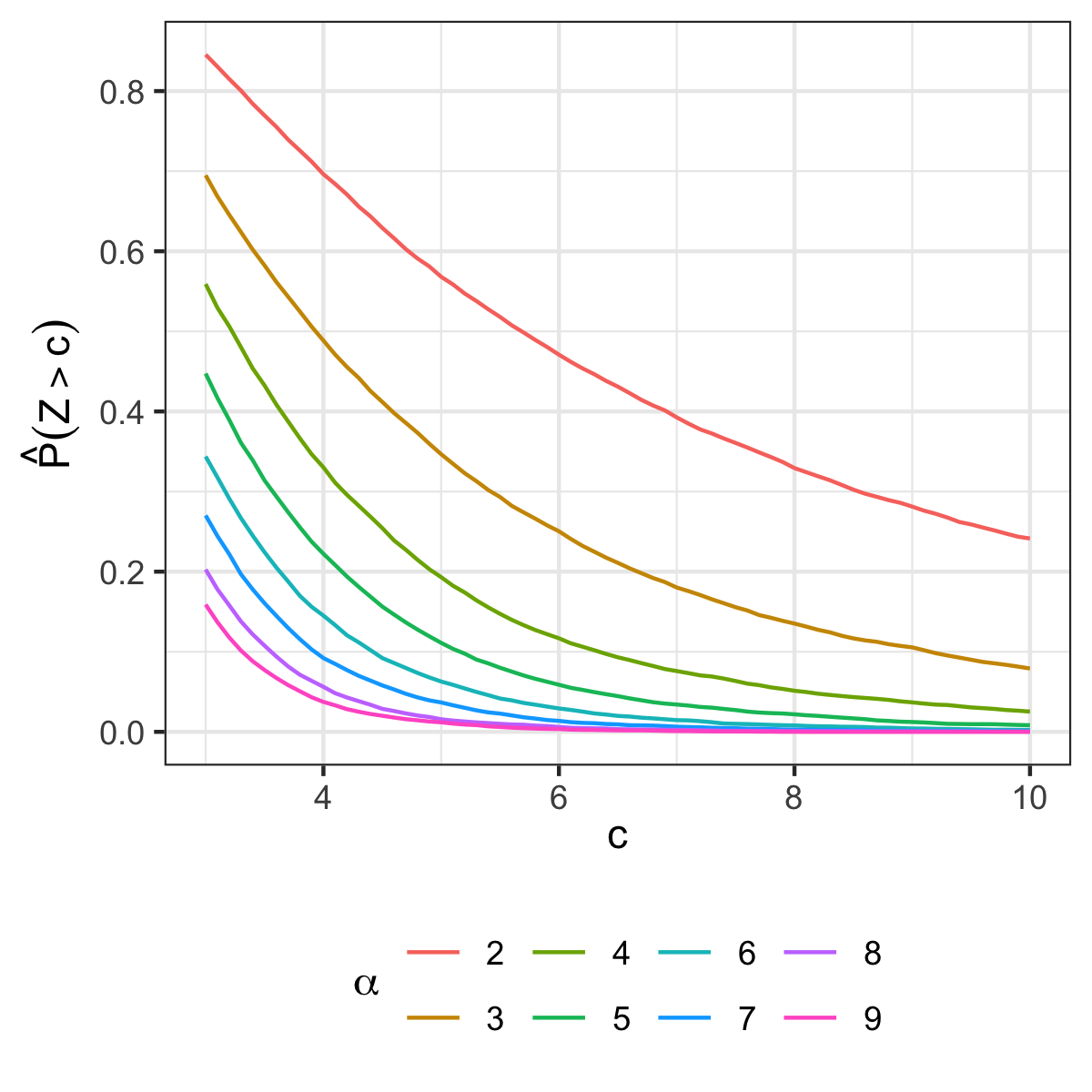}}
\subfigure[k=10]{\includegraphics[width=0.49\textwidth]{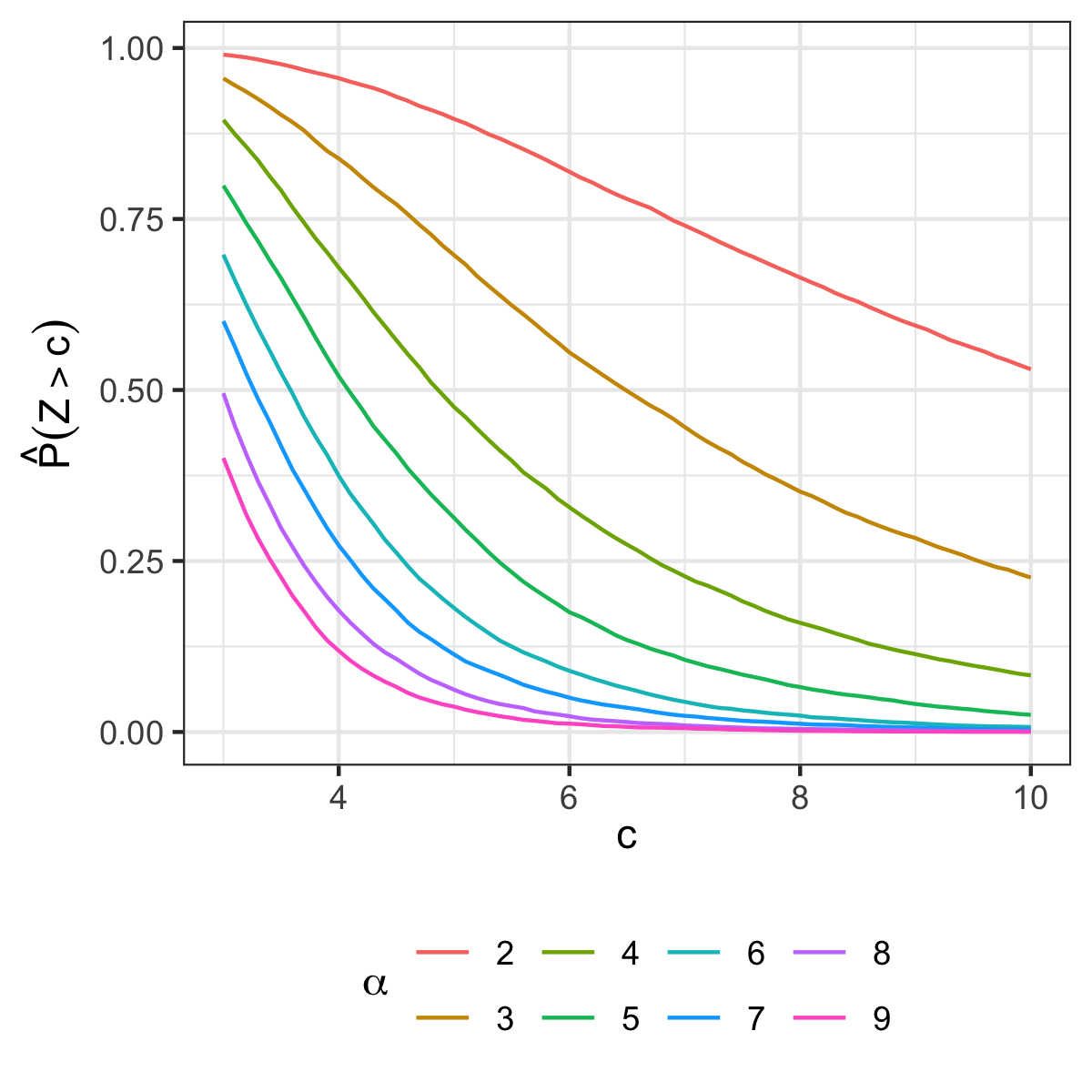}}
\caption[$\alpha$ impact on class-skewness]{\textbf{$\mathbf{\alpha}$ impact on class-skewness}}
\label{fig:Dir_visu}
\end{figure}

As it is indicated by Figure~\ref{fig:Dir_visu}, the lower the $\alpha$ is, the higher is the ratio between majority and minority class(and thus the higher is the class-skewness). The skewness is also notably higher when the number of classes increase ($k=10$). 
To illustrate the skewness impact across different quality of the classifiers, we also 
experimented on three classifiers with parameter $d= 2,5,9$ to represent classifiers of different level of performance. That is, for each of the three classifiers with different covariate availability ($d = 2,5,9$), we sample disproportionately according to Dirichlet-generated weight vectors $\vec{w}\sim \text{Dir}(\alpha)$. The skewness parameter $\alpha$ is varied among $2,5,9$, based upon which, the sampling is repeated 30 times to create $3\times 30 = 90$ imbalanced testing sets for classifier performance evaluation.

To have a glimpse about the weight specification effect, we provided the visualization plots for both "weighted" and "unweighted" ROC-DMF. In the "unweighted" ROC-DMF plot, the factorization cost is universally equal by having $Q_{j} =\frac{1}{W_{j}}, W_j = n_0(j) \times n_1(j), \forall j = 1,\ldots, k(k-1)$ with with $n_0(j), n_1(j)$ defined in Eq~\eqref{eq:binarycard}. In the "weighted" ROC-DMF, we applied factorization weight according to the cardinality (or prior occurrence probability) of the true labels, i.e. $\forall j = 1,\ldots, k(k-1), Q_j = 1, W_{j} = n_0 \times n_1$.
\begin{figure}[H]
\centering
\subfigure[Weighted ROC]{\includegraphics[width=0.49\textwidth]{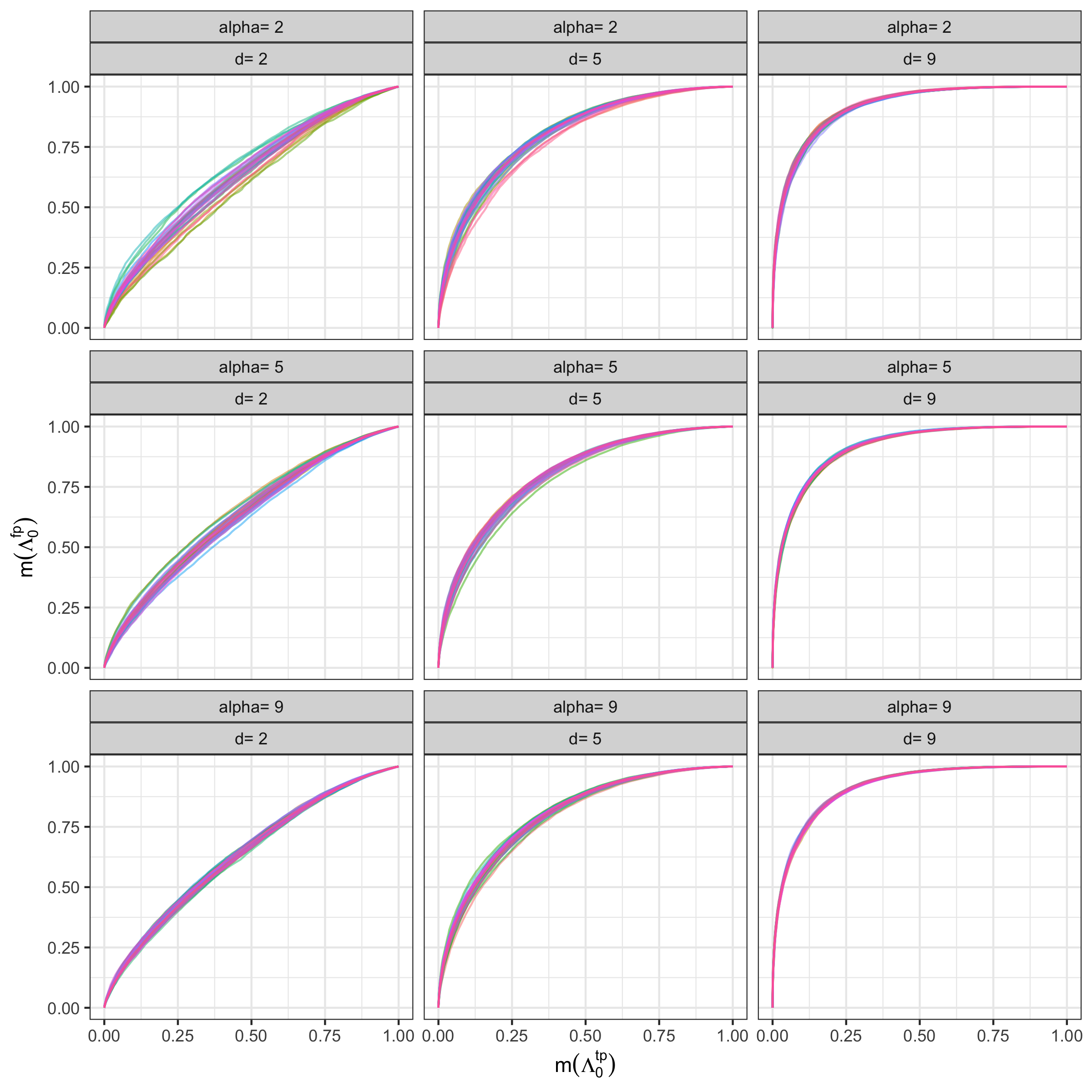}}
\subfigure[Unweighted ROC]{\includegraphics[width=0.49\textwidth]{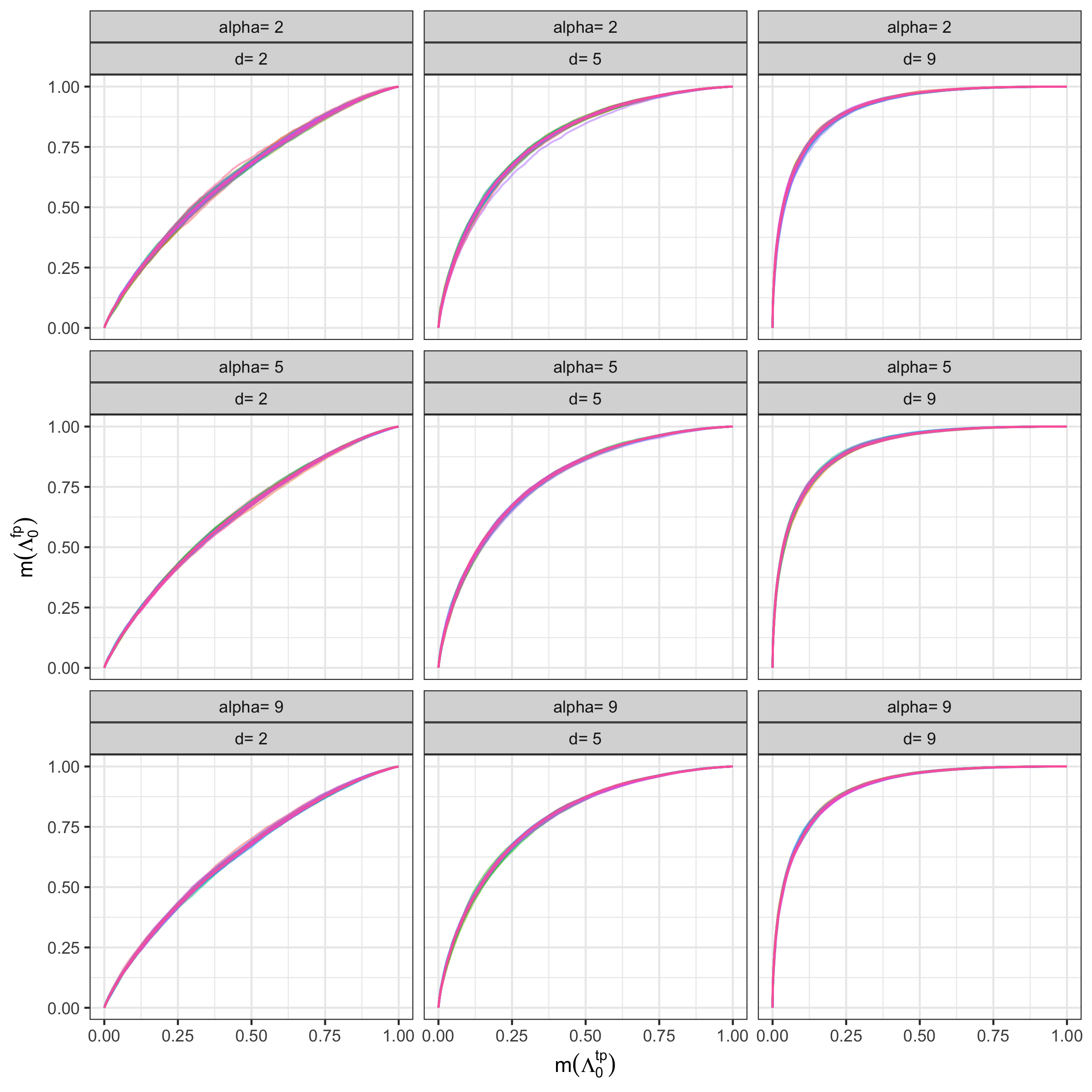}}
\caption[Class-skewness experiment]{\textbf{Class-skewness experiment}}
\label{fig:ClassSkew}
\end{figure}%

From Figure \ref{fig:ClassSkew}, we can conclude that for a given level of classifier defined by the availability covariate dimension $d$, the unweighted ROC curve barely varies. Because the weighted ROC result is obtained through prior probability weighted factorization, there are more variation associated with the low-quality classifiers. Comparing the curves row-wisely according to different skewness level, we can conclude that our constructed ROC curve is invariant to class-skewness. To amplify the weight specification effect, we continue our experiments in the next section.

\subsection{Weight Specification Experiment}
\label{subsec:ResultWeight}
To examine the impact of weights specification on the ROC plot, we created a naive soft classifier that universally predicts the majority-class with dominated probability. In this naive classifier,  all observations universally have the highest probability of belonging to the majority class (e.g. class four from our original 50,000 examples). With equal misclassification cost, the created naive classifier should be considered a bad classifier since it universally predicts all the observations to be the majority class. However, this majority classifier might not be a bad classifier if 
 \begin{enumerate}
    \item the cost of misclassifying the majority class into other classes is low.
    \item the cost of misclassifying other classes into the majority class is high. 
    \item the reward of correctly classifying the majority class is high.
    \item the reward of correctly classifying the other classes is low.
\end{enumerate}
 In fact, naively classifying all the classes into the majority class should be the optimal classifier if the relative weights and reward are extreme. The conclusion will be the opposite if the two cost requirements are exchanged. With the first index pair named as classification class and the second pair index named as reference class according to Figure~\ref{fig:partition}, this cost requirement can be easily specified by the following column-wise operation:
 \begin{enumerate}
     \item decreasing the FPR weight such that the majority class is the classification class.
     \item increasing the FPR weights such that the majority class is the reference class.
     \item increasing the TPR weights such that the majority class is the classification class.
     \item decreasing the TPR weights such that the majority class is  the reference class.
 \end{enumerate}
To test the change of the ROC plot with respect to the change of the misclassification cost $Q_j$, we initialize all the pairwise FPR and pair-wise TPR weights to be one and gradually change the pair-wise TPR/FPR weights associated with the majority class through a multiplication/division on a parameter $c$. 
To illustrate the experiment design, we provide an example of TPR/FPR weight matrix with $k = 3$
 and the first class being the majority class below, that is we choose $Q_j$ such that
\[
\setlength{\arraycolsep}{2pt}
W^{pos} \circ Q^{tp} = \begin{bNiceMatrix}[first-row]
 (1,2) & (1,3) & (2,1) & (2,3) & (3,1) & (3,2)\\
c & c & 1/c & 1 & 1/c & 1 \\
    c & c & 1/c & 1 & 1/c & 1 \\
    $\vdots$ & $\vdots$ & $\vdots$ & $\vdots$ & $\vdots$ & $\vdots$\\
    1 & c & 1/c & 1 & 1/c &1 \\
\end{bNiceMatrix},
W^{neg} \circ Q^{fp}= \begin{bNiceMatrix}[first-row]
 (1,2) & (1,3) & (2,1) & (2,3) & (3,1) & (3,2)\\
1/c & 1/c & c & 1 & c &1 \\
    1/c & 1/c & c & 1 & c &1 \\
    $\vdots$ & $\vdots$ & $\vdots$ & $\vdots$ & $\vdots$ & $\vdots$\\
    1/c & 1/c & c & 1 & c &1 \\
\end{bNiceMatrix}
\]
When $c>1$, it corresponds to the scenario where the relative TPR weight of the majority class is high and the relative FPR weight of the majority class is low. This implies the majority classifier is optimal since the cost of mis-classification is low while the reward the correct classification is high. The scenario is exactly the opposite when $c<1$. The parameter $c$ is then changed from $c = 0.1, c= 0.2, \ldots, c = 0.9$, and $c =1/0.1, c= 1/0,2, \ldots, c= 1/0.9$. 

To demonstrate that this conclusion is invariant to different number of the classes, we choose $n= 10,000$ and vary $k$ from $ 5, 10, 15$. To also validate the impact of class skewness, we simulate the 10,000 classification labels disproportionately according to direction distributed vector $\vec{w} \sim \text{Dir}(\alpha)$ with $\alpha = 2,5,9$. The result is summarized below:
\begin{figure}[H]
\centering
\includegraphics[width=\textwidth, height = \textwidth]{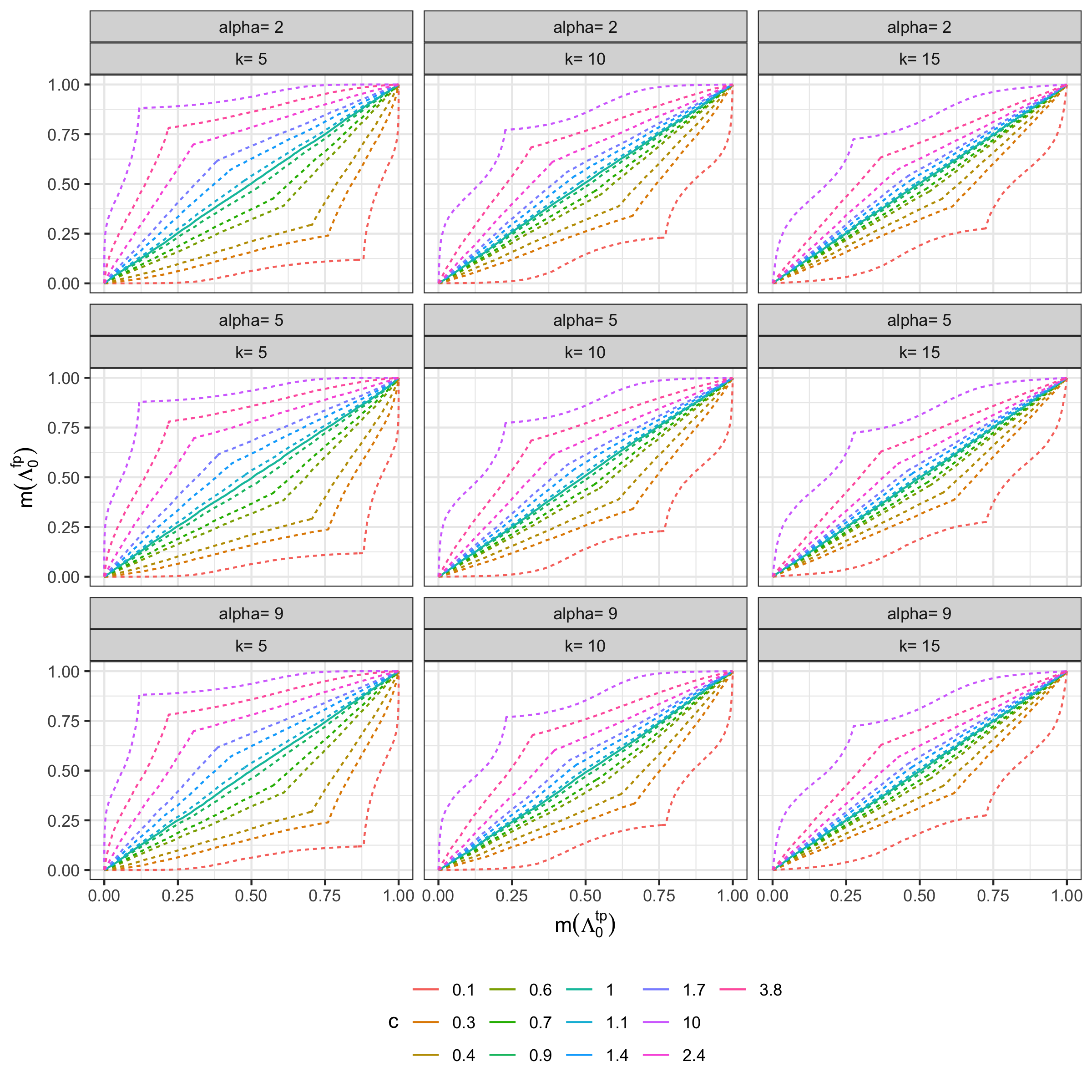}
\caption[Weight specification experiment]{\textbf{Weight specification experiment}}
\label{fig:ClassWeight}
\end{figure}%
As we can see from Figure \ref{fig:ClassWeight}, initially with $c= 1$, the majority classifier can be considered as a random classifier by having the ROC curve as a straight line along the diagonal. The weighted ROC curve then correctly reflected the weight specification by moving the ROC curve to the top-left if $c>1 (c\rightarrow \infty)$ and by moving the ROC curve to the bottom-right if $c<1  (c\rightarrow 0)$.
\subsection{Real World Experiment}
\label{subsec:ResultReal}
To comprehensively examine our methods, we experimented with benchmark datasets that have been used by existing literature. To illustrate the invariance to class-skewness property, \citep{hand2001simple} experimented on eight commonly used multi-class datasets from the UCI Repository of Machine Learning Databases\footnote{https://archive.ics.uci.edu/ml/index.php}. For a comparison to the pair-wise AUC statistics (called $\mathcal{M}$ measurement in \citep{hand2001simple}), we applied our method on the same datasets and adopted the same sets of classification models with similar training procedures. Specifically, for each of the eight datasets, we fit logistic regression, k-nearest neighbor, and decision tree used in Splus\citep{venables1999modern}. 

We adopted the same training procedure by splitting the dataset into equal sizes of training and testing. It is also important to realize that due to the absence of a random seed number,  the ranking of the three model performances can sometimes be different from the Hands' paper due to train/test split. We thus instead focus on the ranking discrepancies among different classification models between our statistics and the $\mathcal{M}$ statistics conditional on our train/test split. To plot the confidence interval of our models, we bootstrap with binomial weight $W_j = n_0(j) \times n_1(j)$ 100 times the computed AUC and the plotted ROC after obtaining the estimation of unweighted $\eta^{tp}$ and $\eta^{fp}$ with $Q_j\circ W_j = 1$. The bootstrapped $\mathbb{M}^{tp}$ and $\mathbb{M}^{fp}$ can thus provide a confidence interval on the estimated $\mathcal{D}$ statistics. We summarize the bootstrapped $\mathcal{D}$ statistics for the six datasets with the following box plot. The pair-wise AUC score is plotted as a blue dot for comparison:
\begin{figure}[H]
\centering
\includegraphics[width=\textwidth]{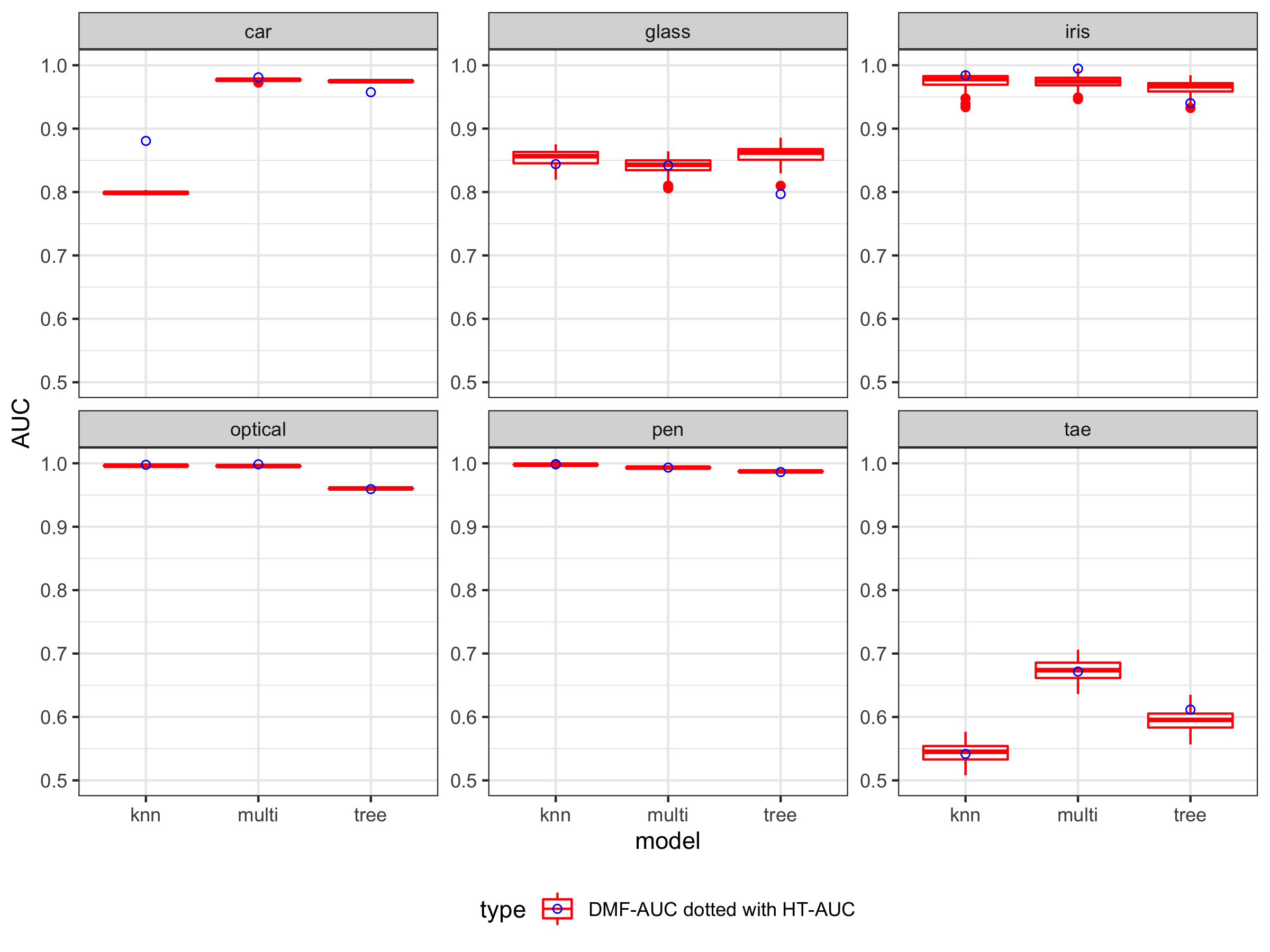}
\caption[AUC comparison]{\textbf{AUC comparison}}
\label{fig:CompAUC}
\end{figure}%
As it is indicated in Figure \ref{fig:CompAUC}, even though the exact AUC numbers obtained from $\mathcal{M}$ and $\mathcal{D}$ are different. The ranking of the model performance tends to agree. Moreover, for some cases when the $\mathcal{M}$ statistics give a performance score with a negligible difference (e.g the IRIS dataset and the GLASS dataset), our $\mathcal{D}$ method quantifies the uncertainty behind the AUC statistics with a confidence interval. This observation additionally substantiates the superiority of our method by having our evaluation metric being able to provide more discriminative ranking when the $\mathcal{M}$ statistics fails to do so \citep{ling2003auc,halimu2019empirical}. In fact, due to the bootstrap sampling of the confidence interval, one can readily estimate the probability of ranking statistics such as:
$\mathbb{P} (\text{AUC}(tree)> \text{AUC}(multinomial) > \text{AUC}(knn))$.
There are six possible permutation of the ranking statistics and we provide their corresponding estimate in the following table:
\begin{table}[H]
\setlength{\tabcolsep}{6pt}
\centering
\resizebox{0.95\textwidth}{!}{
\begin{tabular}{ccccccc}\toprule
Dataset   & knn$>$tree$>$multi   & knn$>$multi$>$tree  & multi$>$tree$>$knn   & multi$>$knn>tree   & tree$>$knn$>$multi   & tree$>$multi$>$knn  \\
\midrule
car     & -    & -    & 0.84 & -    & -    & 0.16 \\
optical & -    & 0.94 & -    & 0.06 & -    & -    \\
pen     & -    & 1.00    & -    & -    & -    & -    \\
tae     & -    & -    & 0.99 & 0.01 & -    & -    \\
iris    & 0.16 & 0.36 & 0.15 & 0.22 & 0.04 & 0.07 \\
glass   & 0.25 & 0.13 & -    & 0.04 & 0.37 & 0.21 \\
\bottomrule
\end{tabular}
}
\caption{Probability estimation of classification model ranking}
\label{tab:pvalue_ranking}
\end{table}

As we can see from Table~\ref{tab:pvalue_ranking}, the car, optical, pen and tae dataset tend to have more certainty in the ultimate classification ranking while it seems that more possible ranking orders are available for the Glass and IRIS dataset. One can also check on the shared threshold effect (ROC) by plotting the $m(\Lambda_0^{tp})$ against $m(\Lambda_0^{fp})$:
\begin{figure}[H]
\centering
\includegraphics[width=\textwidth]{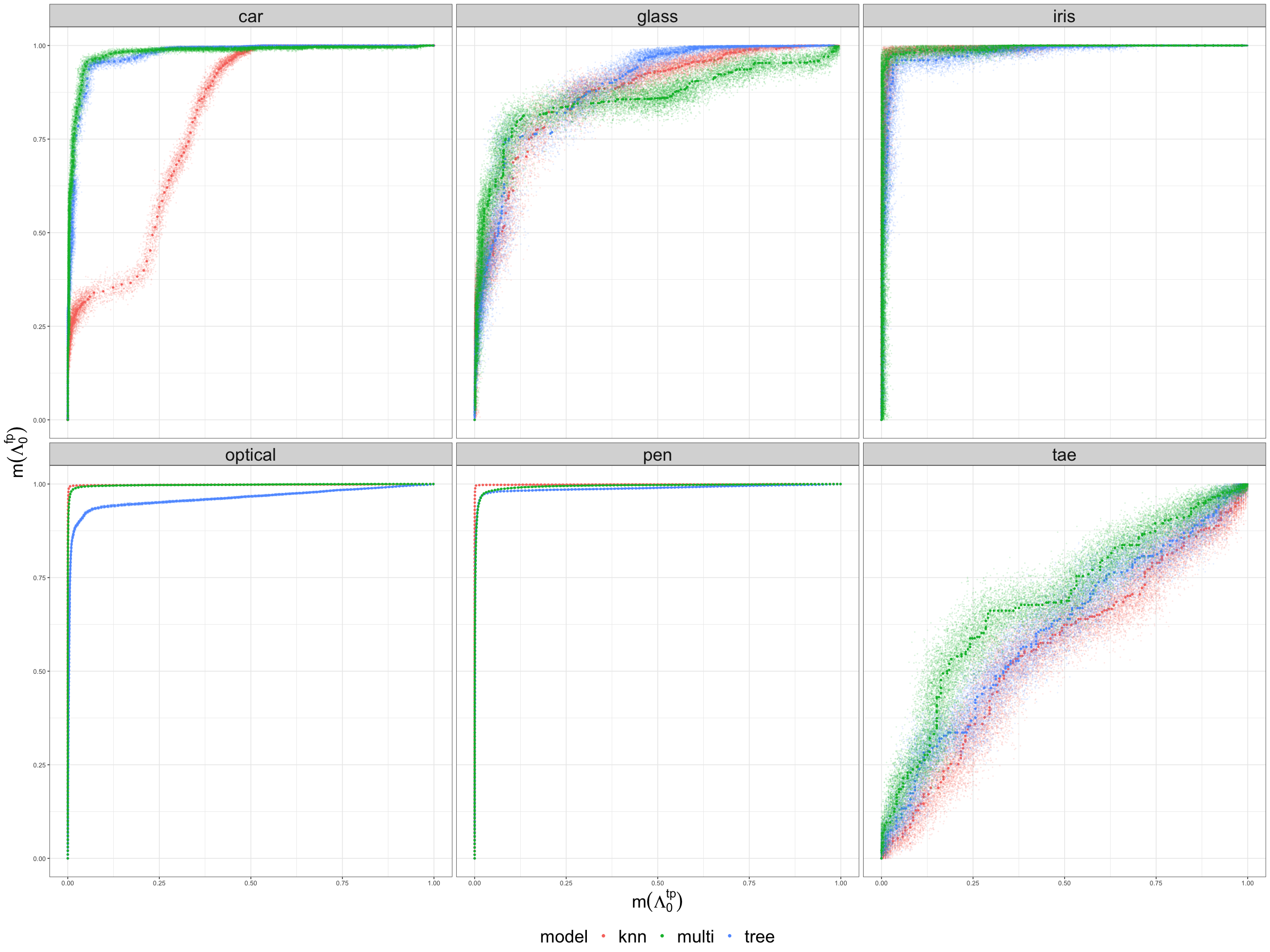}
\caption[Multiclass ROC with confidence interval]{\textbf{Multiclass ROC with confidence interval}}
\label{fig:CompROC}
\end{figure}
From the AUC comparison of the GLASS dataset, we found that the multinomial regression and the KNN have similar AUC scores (0.8419 and 0.8440). Instead of naively concluding that KNN is preferred, a more careful conclusion can be drawn from the ROC plot in Figure~\ref{fig:CompROC}. Specifically, we can find that the multinomial regression model tends to have the highest TPR across all $k(k-1)$ pairs with a high threshold. This can sometimes be preferred by applications (e.g. Alexa Question \& Answering) that prefer a more confident label assignment by thresholding the prediction probability.

\section{Conclusion}
\label{sec:end}
In this paper, we studied an evaluation method for multi-class classification models. The method visually summarize the classifier performance through a rank one factorization. The factorized components coincides with the sROC interpretation by having the column correlation being modeled as random effect. The evaluation method is not only invariant to class skewness but also supports a weight specification for mis-classification loss. A bootstrapped confidence interval is also available to quantify the uncertainty behind the evaluation.

\appendix
\bibliography{references}  

\end{document}